\documentclass{article} 
\usepackage{iclr2027_conference,times}

\usepackage{amsmath,amsfonts,bm}

\def\eqref#1{equation~\ref{#1}}

\def\1{\bm{1}}

\DeclareMathAlphabet{\mathsfit}{\encodingdefault}{\sfdefault}{m}{sl}
\SetMathAlphabet{\mathsfit}{bold}{\encodingdefault}{\sfdefault}{bx}{n}

\usepackage{hyperref}
\usepackage{url}
\usepackage{xspace}
\usepackage{graphicx}
\usepackage{booktabs}
\usepackage{multirow}
\usepackage[normalem]{ulem}
\usepackage[table]{xcolor} 
\usepackage{wrapfig}
\usepackage{enumitem}
\usepackage[most]{tcolorbox}
\usepackage{subcaption}
\usepackage{makecell}

\newcommand{\framework}{\textsc{RoP}\xspace}

\title{Reward on Path: Learning Intermediate Supervision Signals for Knowledge Graph Question Answering}

\author{
Shengxiang Gao$^{1}$, Chao Lei$^{1}$, Jey Han Lau$^{1}$, 
Linhao Luo$^{2}$, Jianzhong Qi$^{1}$\thanks{Corresponding author.} \\
$^{1}$University of Melbourne
\qquad
$^{2}$Monash University \\
\small\texttt{\{shengxiang,clei1\}@student.unimelb.edu.au} \\
\small\texttt{\{jeyhan.lau,jianzhong.qi\}@unimelb.edu.au}
\qquad
\small\texttt{Linhao.Luo@monash.edu}
}

\iclrfinalcopy 
\begin{document}

\maketitle
\fancyhead{}
\lhead{Preprint}

\begin{abstract}
Knowledge Graph Question Answering (KGQA) aims to answer user questions by reasoning over Knowledge Graphs (KGs). Recent methods use supervision derived from answer labels or refined by Large Language Models (LLMs) to train models that retrieve KG evidence for LLM-based answer reasoning. However, answer-derived supervision treats every answer-reaching path as correct and thus yields noisy training signals, whereas LLM-refined supervision mitigates this noise at substantial cost. To address these limitations, we propose Reward on Path (\framework), a framework to  learn a lightweight, question-conditioned path reward from answer labels with an asymmetric objective. Paths reaching the same answer are supervised jointly as a bag, allowing the reward model to learn their relative contributions, while each path is penalized individually for retrieving non-answer entities. The learned reward then trains an LLM-based relation path generator in two stages: reward distillation transfers reward-induced preferences over candidate paths into the generator, and on-policy optimization with GRPO further refines the policy on self-generated paths. Generated paths are grounded in the KG to retrieve evidence for answer reasoning. Experiments on multiple KGQA datasets show that \framework improves F1 over answer-derived methods by at least 2.6\% while outperforming LLM-refined methods without costly supervision construction.
\end{abstract}

\section{Introduction}\label{sec:intro}
Knowledge Graph Question Answering (KGQA) aims to answer users' natural language questions by reasoning over Knowledge Graphs (KGs)~\citep{lan_complexkbqa_2023, pan_unifyingSurvey_2024}. It provides a user-friendly interface for accessing abundant factual knowledge in structured formats stored in large-scale KGs such as Freebase~\citep{bollacker_freebase_2008}, Wikidata~\citep{vrandecic_wikidata_2014}, and DBpedia~\citep{auer_dbpedia_2007}. Recent KGQA methods augment Large Language Models (LLMs) with retrieved KG 
evidence through KG-based Retrieval-Augmented Generation (KG-RAG) methods.

To obtain relevant evidence, existing KG-RAG methods train models to retrieve KG components directly~\citep{li_subgraphrag_2025, mavromatis_gnnrag_2025} or to generate reasoning paths that guide evidence retrieval~\citep{luo_rog_2024, luo_gcr_2025}.
Training such models typically requires intermediate supervision that identifies relevant KG evidence, such as supporting triples, reasoning paths, or subgraphs. This exposes a fundamental challenge: \textbf{the lack of reliable intermediate supervision signals}. Obtaining gold annotations for such evidence is time- and resource-intensive, making reliable intermediate supervision difficult to obtain at scale~\citep{gu_beyond_2021, li_kbBinder_2023, li_flexkbqa_2024}.

\begin{figure}[t]
    \centering
    \includegraphics[width=1\linewidth]{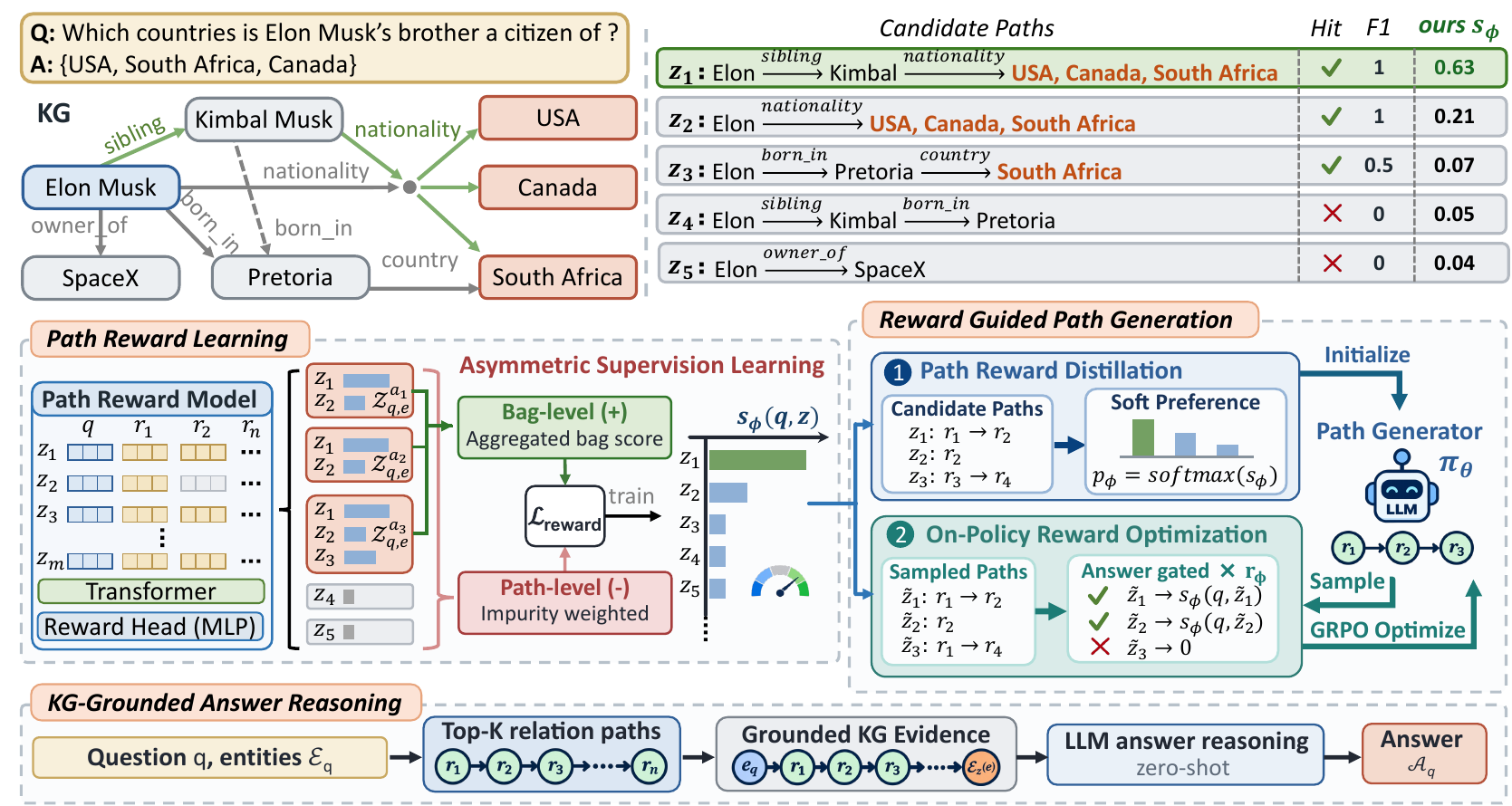}
    \caption{Motivating example and overview of \framework. Top: an illustrative KG subgraph shows that several relation paths reach the correct answers, but only one of them reflects the relations expressed in the question. Bottom: \framework consists of three stages: (1) path reward learning from answer labels, (2) reward-guided path generation through path reward distillation and on-policy reward optimization, and (3) KG-grounded answer reasoning over evidence retrieved by the generated paths.}
    \label{fig:framework}
\end{figure}

One line of work derives intermediate supervision directly from answer labels~\citep{li_subgraphrag_2025, luo_rog_2024, luo_gcr_2025, ma_dp_2025, lu_double_2026}. Reasoning paths that reach ground-truth answers are typically treated as positive training signals. While scalable, this criterion can retain \emph{spurious} paths that reach the correct answer without reflecting the reasoning process required by the question~\citep{lin_kgqarl_2018, yao_rapl_2025}. For example, in Figure~\ref{fig:framework}, both relation paths ``\textit{nationality}'' and ``$\textit{sibling} \rightarrow \textit{nationality}$'' reach the correct answers, although the former does not fully reflect the relations expressed in the question. Consequently, treating all answer-reaching paths as positive can introduce noisy supervision for training retrieval models, thereby degrading evidence retrieval and downstream answer reasoning.

Another line of work employs LLMs to construct or refine intermediate supervision. For example, LLMs can be used to select relevant paths from answer-reaching paths or assign rewards to generated paths~\citep{yao_rapl_2025, zou_reg_2025, wang_damr_2026, luo_kbqao1_2025, wang_kg-hopper_2026}. Although LLM-refined supervision can reduce noise in answer-derived signals and provide more informative intermediate supervision, constructing such supervision typically requires repeated LLM inference over training instances, leading to substantial computational cost and limiting scalability.

To this end, we propose a framework called \textbf{R}eward \textbf{o}n \textbf{P}ath (\framework), which learns path rewards from answer labels to train an LLM-based relation path generator. \framework trains a lightweight question-conditioned reward model with an asymmetric objective that combines bag-level positive supervision with path-level penalties. Intuitively, the bag-level objective allows the reward model to learn and score each candidate path, \textit{without} making the assumption that all answer-reaching paths are equally valid like prior methods. The path-level penalty, on the other hand, penalizes a path if it produces answers that are not part of the ground-truth answer set. Together, these signals enable the reward model to learn the relative utility of candidate paths for answering the question. 

The learned reward model provides intermediate supervision for path generation training in two stages. \emph{Path reward distillation} aligns the generator with the soft preferences induced by the learned rewards over candidate paths. \emph{On-policy reward optimization} applies the same learned reward to self-generated paths and updates the policy through Group Relative Policy Optimization (GRPO). 
Intuitively, these two training schemes are a variation of the standard supervised fine-tuning (SFT) and GRPO training, where we use the learned reward to derive soft targets over candidate paths for the former and to score generated paths in the latter. The resulting path generator produces relation paths that guide KG evidence retrieval, providing grounded context for downstream answer reasoning. The main contributions of this work are summarized as follows:

\begin{itemize}[leftmargin=*]

\item We propose \framework, which learns question-conditioned path rewards from answer labels using an asymmetric objective that combines bag-level positive supervision and path-level penalties.
\item We develop a two-stage training scheme that applies the learned reward to reward distillation and on-policy optimization, extending reward guidance from candidate paths to self-generated paths.
\item Experiments on multiple datasets show that \framework improves F1 over answer-derived methods by at least 2.6\%, while outperforming LLM-refined methods. Further retrieval, ablation, and supervision quality analyses validate the learned path rewards and the two-stage training strategy.

\end{itemize}

\section{Related Work}\label{sec:related_work}

\textbf{Knowledge Graph Question Answering (KGQA).} 
Recent KGQA methods combine LLMs with structured KG evidence through different retrieval and reasoning strategies. KG-RAG methods either employ graph retrievers to directly identify relevant KG components~\citep{li_subgraphrag_2025, mavromatis_gnnrag_2025}, or generate relation paths that serve as retrieval plans for subsequently grounding relevant evidence from the KG~\citep{luo_rog_2024, luo_gcr_2025, ma_dp_2025}. Agent-based methods iteratively interact with the KG, dynamically exploring relevant entities and relations through multi-step reasoning and graph traversal~\citep{sun_tog_2024, chen_pog_2024, luo_kbqao1_2025, wang_damr_2026}. Our work follows the KG-RAG paradigm with relation path generation and focuses on learning reliable intermediate supervision used to train the generator.

\textbf{Intermediate Supervision for KGQA.} Existing methods obtain intermediate supervision either from answer labels or from LLM feedback. The former line treats every relation path connecting a question entity to an annotated answer as a positive training instance~\citep{li_subgraphrag_2025, luo_rog_2024, luo_gcr_2025, ma_dp_2025}; RPO-RAG~\citep{um_rporag_2026} further ranks such answer-reaching paths by their similarity to the question under a pretrained encoder. The latter line invokes LLMs at training time: RAPL~\citep{yao_rapl_2025} and ReG~\citep{zou_reg_2025} use LLM feedback to filter spurious paths, DAMR~\citep{wang_damr_2026} and KBQA-o1~\citep{luo_kbqao1_2025} use LLM-guided search to construct or refine reasoning trajectories, and KG-Hopper~\citep{wang_kg-hopper_2026} uses an LLM to reward generated paths during GRPO training.

\framework also learns from answer labels but differs from answer-derived approaches in how intermediate supervision is derived and used. It learns a question-conditioned path reward with an asymmetric objective to estimate the relative utility of candidate paths, and uses it to supervise both path reward distillation and on-policy reward optimization, reducing supervision noise without LLM feedback.

\section{Preliminary}\label{sec:preliminary}

\textbf{Knowledge Graphs (KGs)} are composed of relational facts in the form of a set of triples:
$\mathcal{G} = \{ \langle e_h, r, e_t \rangle \mid e_h \in \mathcal{E}, r \in \mathcal{R}, e_t \in \mathcal{E}\}$,
where $\mathcal{E}$ denotes the set of entities and $\mathcal{R}$ denotes the set of relations. In each triple $\langle e_h, r, e_t \rangle$, $e_h,e_t \in \mathcal{E}$ denote the head and tail entities, respectively, and $r \in \mathcal{R}$ represents the relation between them.

\textbf{Knowledge Graph Question Answering (KGQA)} is a reasoning task over KGs. Given a KG $\mathcal{G}$ and a question $q=(w_1,w_2,\ldots,w_n)$ with linked entities $\mathcal{E}_q\subseteq\mathcal{E}$, where $w_i$ is the $i$-th token, the objective is to learn a function $f$ that maps $(q,\mathcal{E}_q,\mathcal{G})$ to an answer set $\mathcal{A}_q\subseteq\mathcal{E}$, i.e., $\mathcal{A}_q=f(q,\mathcal{E}_q,\mathcal{G})$.

\textbf{Relation Path} is an ordered sequence of relations
$z=\langle r_1,r_2,\ldots,r_l \rangle$, where $r_i \in \mathcal{R}$ denotes the $i$-th relation and $l$ denotes the path length. 
It defines a compositional relation that connects a source entity to a set of target entities by following the relations in order (example provided in Appendix~\ref{app:path_example}). 
Given an entity $e$ and a relation path $z$, we denote the set of entities reachable from $e$ via $z$:
\[
\mathcal{E}_z(e)
=
\left\{
e_l \in \mathcal{E}
\mid
e \xrightarrow{r_1} e_1 \xrightarrow{r_2} \cdots \xrightarrow{r_l} e_l
\text{ in } \mathcal{G}
\right\}.
\]
\section{Methodology}\label{sec:methodology}

As illustrated in Figure~\ref{fig:framework}, \framework consists of three main stages: (1) \emph{path reward learning}, which learns a lightweight question-conditioned reward model $s_\phi$ from answer labels; (2) \emph{reward-guided relation path generation}, which uses the learned reward to train an LLM-based relation path generator $\pi_\theta$ through path reward distillation and on-policy reward optimization; and (3) \emph{KG-grounded answer reasoning}, which grounds the generated paths in the KG to retrieve evidence and employs an LLM answer generator $p_\psi$ to answer the question via zero-shot prompting.

\subsection{Path Reward Learning}\label{subsec:reward}

To obtain reliable intermediate supervision, we learn a lightweight path reward model from answer labels to estimate the utility of candidate relation paths for answering the question. Formally, given a question $q$ and a candidate relation path $z$, the model assigns a scalar reward score $s_\phi(q,z)\in\mathbb R$.

For each question entity $e\in\mathcal{E}_q$, we construct a set $\mathcal{Z}_{q,e}=\{z_i\}_{i=1}^{N}$ by breadth-first search from $e$ up to $L$ hops, retaining all answer-reaching paths and randomly sampling from the remaining paths. We exclude question--entity pairs with no answer-reaching candidate paths. Each path $z_i$ retrieves a set of end entities $\mathcal{E}_{z_i}(e)$ when grounded in the KG. Sampling details are provided in Appendix~\ref{app:implementation}.

\textbf{Path Reward Modeling.} We parameterize $s_\phi(q,z)$ with a lightweight Transformer encoder that captures question-dependent path utility by jointly encoding the question and the relation sequence. For a path $z=\langle r_1,r_2,\ldots,r_l\rangle$, a frozen text encoder produces a question embedding $\mathbf{h}_q\in\mathbb{R}^d$ and relation embeddings $\{\mathbf{h}_{r_i}\}_{i=1}^{l}$. We add learnable positional embeddings $\{\mathbf{p}_j\}_{j=0}^{l}$ and feed the resulting sequence into the Transformer. The output at the question position serves as a question-conditioned path representation, which an MLP reward head maps to a scalar reward score:
\begin{equation}
\mathbf{H}_{q,z}
=
\mathrm{Transformer}_{\phi}\!\left(
[\mathbf{h}_q+\mathbf{p}_0,\,
 \mathbf{h}_{r_1}+\mathbf{p}_1,\,
 \ldots,\,
 \mathbf{h}_{r_l}+\mathbf{p}_l]
\right),
\qquad
s_\phi(q,z)
=
\mathrm{MLP}_{\phi}\!\left(\mathbf{H}_{q,z}^{(0)}\right).
\label{eq:path_score}
\end{equation}
The trainable parameters $\phi$ include the Transformer encoder, positional embeddings, and MLP, while the text encoder remains frozen.

\textbf{Asymmetric Reward Learning.} Paths reaching the same answer may not all reflect the relations expressed in the question. We therefore group these paths into an answer-specific bag and supervise its aggregate prediction rather than each path individually, allowing the reward model to learn their relative contributions. For each annotated answer $a\in\mathcal{A}_q$, we define
$\mathcal{Z}_{q,e}^{a}=\{z_i\in\mathcal{Z}_{q,e}\mid
a\in\mathcal{E}_{z_i}(e)\}$ and denote the answers with nonempty bags by
$\mathcal{A}_{q,e}^{+}=\{a\in\mathcal{A}_q\mid
\mathcal{Z}_{q,e}^{a}\neq\emptyset\}$.

For each $a\in\mathcal{A}_{q,e}^{+}$, the reward scores define attention
weights over the paths in its bag, and the bag prediction is the
attention-weighted average of the individual path predictions:
\begin{equation}
\omega_{q,e}^{a}(z_i)
=
\frac{
\exp\left(s_\phi(q,z_i)\right)
}{
\sum_{z_j\in\mathcal{Z}_{q,e}^{a}}
\exp\left(s_\phi(q,z_j)\right)
},
\qquad
\hat{y}_{q,e}^{a}
=
\sum_{z_i\in\mathcal{Z}_{q,e}^{a}}
\omega_{q,e}^{a}(z_i)
\,\sigma\left(s_\phi(q,z_i)\right).
\end{equation}
Here, $\sigma(\cdot)$ denotes the sigmoid function, and
$\sigma(s_\phi(q,z_i))$ is the individual prediction for path $z_i$.

To complement this aggregate supervision, we penalize individual paths according to their \emph{answer impurity}, defined as the fraction of retrieved entities outside the annotated answer set:
\begin{equation}
\xi_{z_i}
=
1-
\frac{
\left|\mathcal{E}_{z_i}(e)\cap\mathcal{A}_q\right|
}{
\left|\mathcal{E}_{z_i}(e)\right|
}.
\end{equation}
We use answer impurity to weight the penalties on individual paths and combine them with positive supervision on answer-specific bags, yielding the following asymmetric objective:
\begin{equation}
\mathcal{L}_{\mathrm{score}}
=
\underbrace{
-\frac{1}{|\mathcal{A}_{q,e}^{+}|}
\sum_{a\in\mathcal{A}_{q,e}^{+}}
\log \hat{y}_{q,e}^{a}
}_{\text{Bag-level positive supervision}}
-
\underbrace{
\frac{1}{|\mathcal{Z}_{q,e}|}
\sum_{z_i\in\mathcal{Z}_{q,e}}
\xi_{z_i}
\log\left(1-\sigma\left(s_\phi(q,z_i)\right)\right)
}_{\text{Path-level penalty}}.
\label{eq:reward_loss}
\end{equation}
The objective is asymmetric in its supervision granularity: positive supervision applies to answer-specific bags, whereas impurity-weighted penalties apply to individual paths. This formulation avoids treating every answer-reaching path as a positive instance while penalizing the retrieval of non-answer entities. By learning question--path interactions across training examples and the relative contributions of paths within each bag, the shared reward model can effectively distinguish paths with otherwise identical answer-derived supervision and provide more informative rewards for path generation. Section~\ref{sec:reward_analysis} empirically validates the quality of the learned supervision.

\subsection{Reward-Guided Relation Path Generation}\label{subsec:training}

The learned reward model $s_\phi$ provides supervision for training the relation path generator $\pi_\theta$ in two stages and remains frozen throughout. \emph{Path reward distillation} fits the generator to soft preferences induced by the reward model over answer-reaching candidate paths. \emph{On-policy reward optimization} then uses the same model to provide graded feedback on paths sampled from the current policy, guiding further refinement through GRPO. In both stages, the generator is trained to produce a relation path given the question and source entity, following the template in Figure~\ref{fig:path_gen_prompt}.

\textbf{Path Reward Distillation.} For each question--entity pair $(q,e)$, we convert the learned reward scores into a soft target distribution over candidate paths that reach at least one answer:
\begin{equation}
p_{\phi}(z_i\mid q,e)
=
\frac{
\exp\left(s_{\phi}(q,z_i)\right)
}{
\sum_{z_j\in\mathcal{Z}_{q,e}^{\mathcal{A}_q}}
\exp\left(s_{\phi}(q,z_j)\right)
},
\qquad
\text{where}\quad
\mathcal{Z}_{q,e}^{\mathcal{A}_q}
=
\bigcup_{a\in\mathcal{A}_{q,e}^{+}}
\mathcal{Z}_{q,e}^{a}.
\label{eq:distribution}
\end{equation}
Let $\pi_\theta(z\mid q,e)=\prod_{t=1}^{T_z}
\pi_\theta(x_t\mid q,e,x_{<t})$ denote the autoregressive probability
of a relation path $z$, where $(x_1,\ldots,x_{T_z})$ is the token
sequence representing $z$. With $p_\phi$ defined as zero outside the
answer-reaching candidate set, we fit the generator to the
reward-induced target distribution by minimizing the forward
KL divergence:
\begin{equation}
\begin{aligned}
\mathcal{L}_{\mathrm{distill}}
&=
D_{\mathrm{KL}}\!\left(
p_{\phi}(\cdot\mid q,e)
\,\|\,
\pi_{\theta}(\cdot\mid q,e)
\right)\\
&=
-\sum_{z_i\in\mathcal{Z}_{q,e}^{\mathcal{A}_q}}
p_{\phi}(z_i\mid q,e)
\log \pi_{\theta}(z_i\mid q,e)
+\mathrm{const.}
\end{aligned}
\label{eq:reward_distill}
\end{equation}
Here, the constant is the negative entropy of $p_\phi$, which is independent of $\theta$ and can therefore be omitted during optimization. Path reward distillation trains the generator on relation sequences extracted from the KG, weighting each sequence by its reward-induced probability rather than uniformly as in standard SFT over answer-reaching candidates. This provides a supervised initialization for subsequent GRPO training that incorporates both KG relation patterns and learned path preferences.

\textbf{On-Policy Reward Optimization.} Starting from the distilled generator, we further optimize path generation with Group Relative Policy Optimization (GRPO)~\citep{shao_grpo_2024}. We construct the reward by scaling a binary answer-reaching signal using the same learned reward model, providing graded feedback on generated paths. For each question--entity pair $(q,e)$, we sample a group of relation paths $\{\tilde{z}_i\}_{i=1}^{G}$ using the old policy $\pi_{\theta_{\mathrm{old}}}$ and assign each path the following reward:
\begin{equation}
u_i
=
\mathbb{I}
\left[
\mathcal{E}_{\tilde{z}_i}(e)
\cap
\mathcal{A}_q
\neq
\emptyset
\right]
\sigma\left(s_\phi(q,\tilde{z}_i)\right),
\label{eq:gated_path_reward}
\end{equation}
where $\mathbb{I}[\cdot]$ denotes the indicator function. It restricts positive rewards to paths reaching an annotated answer, while the learned scores provide graded feedback on their utility for answering the question.

Let $\tilde{x}_{i,t}$ denote the $t$-th token of the generated path $\tilde{z}_i$. Following GRPO, we normalize rewards within each group to obtain relative advantages and compute token probability ratios between the updated and old policies as follows:
\begin{equation}
\hat{A}_i
=
\frac{
u_i-\operatorname{mean}\left(\{u_j\}_{j=1}^{G}\right)
}{
\operatorname{std}\left(\{u_j\}_{j=1}^{G}\right)
+\epsilon_{\mathrm{std}}
},
\qquad
\rho_{i,t}(\theta)
=
\frac{
\pi_\theta(\tilde{x}_{i,t}\mid q,e,\tilde{x}_{i,<t})
}{
\pi_{\theta_{\mathrm{old}}}
(\tilde{x}_{i,t}\mid q,e,\tilde{x}_{i,<t})
}.
\end{equation}
The generator is then updated by minimizing the clipped GRPO objective without KL regularization:
\begin{equation}
\mathcal{L}_{\mathrm{GRPO}}
=
-
\mathbb{E}_{(i,t)}
\left[
\min
\left(
\rho_{i,t}(\theta)\hat{A}_i,\,
\operatorname{clip}
\left(
\rho_{i,t}(\theta),
1-\varepsilon,
1+\varepsilon
\right)
\hat{A}_i
\right)
\right],
\label{eq:grpo_objective}
\end{equation}
where the expectation averages over all generated tokens in the training batch, $\epsilon_{\mathrm{std}}>0$ ensures numerical stability when normalizing group rewards, and $\varepsilon$ controls the clipping range of the policy probability ratio. Sampling settings and optimization hyperparameters are detailed in Appendix~\ref{app:implementation}.

Overall, the two stages apply the same learned reward to different path sources: path reward distillation weights KG-extracted candidate paths by their reward-induced probabilities to initialize the generator, and on-policy reward optimization scores self-generated paths to refine it. The learned reward thus guides both the initialization over candidate paths and the subsequent refinement over self-generated ones, prioritizing paths by their estimated utility rather than by answer outcomes.

\subsection{KG-Grounded Answer Reasoning}\label{subsec:answergen}

At inference time, the trained generator $\pi_\theta$ produces $K$ candidate relation paths for each question entity $e\in\mathcal{E}_q$ using beam search. We ground each path in the KG from $e$ to retrieve its reachable end entities:
\begin{equation}
\widetilde{\mathcal{Z}}_{q,e}^{K}
=
\operatorname{BeamSearch}_{K}\!\left(
\pi_\theta(\cdot\mid q,e)
\right),
\qquad
c_{e,\tilde{z}}
=
\left(e,\tilde{z},\mathcal{E}_{\tilde{z}}(e)\right),
\quad
\tilde{z}\in\widetilde{\mathcal{Z}}_{q,e}^{K}.
\end{equation}
Each evidence item $c_{e,\tilde{z}}$ contains the source entity, relation sequence, and retrieved end entities. We aggregate these items across question entities and generated paths into an evidence context $\mathcal{C}_q$.

The question $q$ and evidence context $\mathcal{C}_q$ are provided to the LLM answer generator $p_\psi$ through zero-shot prompting, using the template in Figure~\ref{fig:reasoning_prompt}. The model is instructed to select answers from the retrieved end entities and generates an answer sequence $y=(y_1,\ldots,y_T)$ according to 
\begin{equation}
p_\psi(y\mid q,\mathcal{C}_q)
=
\prod_{t=1}^{T}
p_\psi(y_t\mid q,\mathcal{C}_q,y_{<t}).
\label{eq:answer_generation}
\end{equation}
The resulting evidence context preserves question-relevant relational structure in a compact form. The answer generator then uses this context to jointly reason over multiple relation paths and their retrieved candidate answers for answering multi-hop and multi-constraint questions.

\section{Experiments}\label{sec:experiment}
We conduct experiments to answer the following research questions:
\textbf{RQ1:} How does \framework compare with existing KGQA methods in effectiveness and efficiency?
\textbf{RQ2:} How does reward model design affect KGQA performance?
\textbf{RQ3:} Can the learned path rewards improve other retrieval models?
\textbf{RQ4:} How does the two-stage reward-guided training strategy contribute to KGQA performance?
\textbf{RQ5:} How reliable are the learned path rewards as intermediate supervision?
\subsection{Experimental Setup}

\textbf{Datasets.}
We evaluate on WebQuestionsSP (WebQSP)~\citep{yih_webqsp_2016} and ComplexWebQuestions (CWQ)~\citep{talmor_cwq_2018}, using the preprocessed Freebase subgraphs adopted by prior work~\citep{luo_rog_2024,mavromatis_gnnrag_2025,luo_gcr_2025}. WebQSP contains 1- to 2-hop questions from Google search logs with annotated SPARQL queries, while CWQ focuses on compositional and multi-constraint questions of up to 4 hops. To evaluate \framework on a different KG, we additionally report results on MetaQA~\citep{zhang_metaqa_2018} (Appendix~\ref{app:metaqa}). Detailed dataset statistics are provided in Appendix~\ref{app:datasets}.

\textbf{Evaluation Metrics.}
For answer generation, we adopt F1, Hit, and Hits@1 as the evaluation metrics. F1 measures answer set correctness using the harmonic mean of answer precision and recall. Hit assesses whether any ground-truth answer appears in the predicted answer set. Hits@1 measures the proportion of questions for which the top-1 predicted answer is correct. For retrieval, we report SPARQL path Hits@1 and Hit using paths extracted from annotated SPARQL queries as references, together with precision, recall, and F1 for retrieved entities against the annotated answer set. For supervision quality, we rank answer-reaching paths and report SPARQL path Hits@1 and MRR.

\textbf{Baselines.}
We compare \framework against recent KG-RAG baselines grouped by their supervision paradigms: (1) \textit{prompting-based methods}, (2) \textit{training with answer-derived supervision}, and (3) \textit{training with LLM-refined supervision}. Detailed descriptions are provided in Appendix~\ref{app:baselines}.

\textbf{Implementation Details.}
The reward model is implemented by a two-layer Transformer encoder and an MLP scoring head, using gte-large-en-v1.5~\citep{li_gte_2023} as the frozen text encoder for questions and relations. We train the LoRA parameters~\citep{hu_lora_2021} of a LLaMA3.1-8B-Instruct relation path generator, keeping its base weights and the learned reward model frozen. At inference, we generate paths with beam size $K=5$ and use GPT-4o as the default answer LLM. Further configurations and parameter studies are provided in Appendices~\ref{app:implementation} and~\ref{app:param_study}.

\subsection{Main Results (RQ1)}\label{sec:main_result}

\textbf{KGQA Performance.}
We first compare \framework with other KG-RAG baselines on WebQSP and CWQ to evaluate its overall KGQA performance. The results for \framework are averaged over three independent runs, with standard deviations reported in Appendix~\ref{app:stand_deviation}. As shown in Table~\ref{tab:main}, \framework achieves the strongest overall accuracy across the two datasets. Among trained methods, \framework consistently outperforms those using answer-derived supervision, with relative gains of 3.0\% and 2.6\% in F1 and up to 7.6\% in Hits@1 across WebQSP and CWQ. These consistent gains show that more reliable intermediate supervision can improve KGQA accuracy. Although LLM-refined baselines invoke LLMs to enhance intermediate supervision, \framework still achieves higher F1 and Hits@1 on both datasets without requiring additional LLM calls for supervision construction. Overall, these results validate the effectiveness of learning path rewards directly from answer labels for improving KGQA reasoning.

\begin{table}[t]
\centering
\caption{Comparison of KGQA performance (\%) on WebQSP and CWQ. \texttt{Model} denotes the model used for answer generation. We highlight the best performance in \textbf{bold} and \underline{underline} the second-best. $^\dagger$ indicates that the model is fine-tuned. $^*$ indicates results taken from the original paper.}
\vspace{-2mm}
\label{tab:main}
\resizebox{\textwidth}{!}{%
\begin{tabular}{lccccccc}
\toprule
\multirow[b]{2}{*}{\textbf{Method}}
& \multirow[b]{2}{*}{\textbf{Model}}
& \multicolumn{3}{c}{\textbf{WebQSP}}
& \multicolumn{3}{c}{\textbf{CWQ}} \\
\cmidrule(lr){3-5}\cmidrule(lr){6-8}
& & \textbf{F1} & \textbf{Hit} & \textbf{Hits@1}
& \textbf{F1} & \textbf{Hit} & \textbf{Hits@1} \\
\midrule
\rowcolor{gray!12}
\multicolumn{8}{l}{\textbf{\textit{Prompting with in-context learning}}} \\
\midrule
ToG~\citep{sun_tog_2024}
& GPT-4o & 50.9 & 78.5 & 72.1 & 42.3 & 56.9 & 51.8 \\
DoG~\citep{ma_debate_2025}
& GPT-4o & 56.1 & 91.2 & 66.9 & 47.5 & 58.4 & 43.0 \\
FiDeLiS~\citep{sui_fidelis_2025}
& GPT-4o & 54.8 & 77.6 & 70.0 & 46.8 & 58.1 & 50.2 \\
\midrule
\rowcolor{gray!12}
\multicolumn{8}{l}{\textbf{\textit{Training with answer-derived supervision}}} \\
\midrule
RoG~\citep{luo_rog_2024}
& LLaMA2-Chat-7B$^\dagger$
& 67.0 & 81.9 & 78.3 & 53.6 & 59.2 & 53.7 \\
GNN-RAG~\citep{mavromatis_gnnrag_2025}
& LLaMA2-Chat-7B$^\dagger$
& 69.2 & 81.1 & 76.2 & 58.3 & 65.3 & 60.5 \\
GCR~\citep{luo_gcr_2025}
& GPT-4o
& 69.8 & 86.9 & 82.9 & 57.1 & 64.6 & 59.1 \\
SubgraphRAG~\citep{li_subgraphrag_2025}
& GPT-4o
& 76.1 & 89.9 & \uline{84.3} & 59.2 & 67.3 & 60.7 \\
DP~\citep{ma_dp_2025}
& GPT-4o
& 75.8 & 88.9 & 81.6 & 57.3 & 63.2 & 60.1 \\
RPO-RAG~\citep{um_rporag_2026}
& LLaMA3.1-8B-Instruct$^\dagger$
& 79.5 & 88.6 & 80.7 & 61.2 & \uline{69.3} & 60.3 \\
\midrule
\rowcolor{gray!12}
\multicolumn{8}{l}{\textbf{\textit{Training with LLM-refined supervision}}} \\
\midrule
KG-Hopper$^*$~\citep{wang_kg-hopper_2026}
& Qwen-2.5-7B$^\dagger$
& - & - & 83.2 & - & - & \uline{61.1} \\
RAPL$^*$~\citep{yao_rapl_2025}
& GPT-4o
& \uline{80.7} & \textbf{93.3} & - & 58.8 & 69.0 & - \\
ReG$^*$~\citep{zou_reg_2025}
& GPT-4o
& 78.8 & \uline{91.4} & - & \uline{62.5} & 68.9 & - \\
\midrule
\rowcolor{gray!12}
\multicolumn{8}{l}{\textbf{\textit{Ours}}} \\
\midrule
\framework
& GPT-4o
& \textbf{81.9} & 91.0 & \textbf{86.0}
& \textbf{62.8} & \textbf{71.1} & \textbf{65.3} \\
\bottomrule
\end{tabular}%
}
\end{table}

\textbf{Retrieval Performance.}
Table~\ref{tab:retrieval} provides a closer view of the retrieval behavior underlying the KGQA results. For path generation, \framework improves SPARQL path Hits@1 over all baselines by at least 15.4\% on both datasets, while attaining the highest SPARQL path Hit on WebQSP and a competitive Hit on CWQ. For answer retrieval, \framework also achieves the highest F1 and precision on both datasets, outperforming the second-best method, DP, by 11.7\% and 8.2\% in F1 and by 14.5\% and 12.1\% in precision on WebQSP and CWQ, respectively. Although GCR attains higher answer recall and DP a higher SPARQL path Hit on CWQ, \framework yields substantially higher Hits@1, precision, and F1, suggesting that its gains arise from more precise path generation rather than broader retrieval coverage. This pattern is consistent with the learned path rewards mitigating noise in answer-derived supervision and providing more reliable intermediate supervision for path generation.

\begin{table}[tbp]
\centering
\caption{Comparison of retrieval performance (\%) on WebQSP and CWQ.}
\label{tab:retrieval}
\vspace{-2mm}
\resizebox{\textwidth}{!}{
\begin{tabular}{lcccccccccc}
\toprule
\multirow{3}{*}{\textbf{Method}}
& \multicolumn{5}{c}{\textbf{WebQSP}}
& \multicolumn{5}{c}{\textbf{CWQ}} \\
\cmidrule(lr){2-6}\cmidrule(lr){7-11}
& \multicolumn{2}{c}{\textbf{SPARQL Path}}
& \multicolumn{3}{c}{\textbf{Answer}}
& \multicolumn{2}{c}{\textbf{SPARQL Path}}
& \multicolumn{3}{c}{\textbf{Answer}} \\
\cmidrule(lr){2-3}\cmidrule(lr){4-6}
\cmidrule(lr){7-8}\cmidrule(lr){9-11}
& \textbf{Hits@1} & \textbf{Hit} & \textbf{F1}
& \textbf{Recall} & \textbf{Precision}
& \textbf{Hits@1} & \textbf{Hit} & \textbf{F1}
& \textbf{Recall} & \textbf{Precision} \\
\midrule
RoG
& 62.3 & 76.2 & 48.1 & 86.1 & 42.5
& 45.1 & 62.3 & 21.1 & 71.1 & 17.4 \\
GCR
& \uline{72.0} & 87.6 & 42.5 & \textbf{91.9} & 34.7
& \uline{54.4} & 62.7 & 18.9 & \textbf{75.6} & 14.3 \\
SubgraphRAG
& - & - & 10.3 & \uline{90.6} & 7.2
& - & - & 3.7 & \uline{74.0} & 2.0 \\
DP
& 69.3 & \uline{87.7} & \uline{62.5} & 84.3 & \uline{58.5}
& 44.8 & \textbf{81.0} & \uline{26.9} & 73.4 & \uline{22.3} \\
RPO-RAG
& 46.6 & 82.8 & 46.1 & 89.1 & 38.5
& 40.8 & 73.7 & 23.1 & 72.6 & 18.9 \\
\midrule
\framework
& \textbf{83.1} & \textbf{88.7} & \textbf{69.8} & 85.7 & \textbf{67.0}
& \textbf{68.6} & \uline{76.2} & \textbf{29.1} & \uline{74.0} & \textbf{25.0} \\
\bottomrule
\end{tabular}
}
\end{table}

\textbf{Efficiency.}
We further compare inference cost and the cost of constructing training supervision. As shown in Table~\ref{tab:efficiency}, \framework achieves the lowest runtime on both datasets while requiring substantially fewer tokens for answer reasoning. This efficiency follows from the precise path generation observed above. The generated paths yield a compact evidence context that reduces the input processed at inference time and shields the answer LLM from irrelevant evidence that would otherwise mislead its reasoning. For supervision construction, LLM-refined methods incur additional LLM inference over the training set. RAPL, the LLM-refined baseline in Table~\ref{tab:efficiency}, requires one GPT-4o call and over 2K tokens per training question, amounting to roughly 6M and 93M tokens over the WebQSP and CWQ training sets, respectively.\footnote{Approximately \$22 and \$318 under the GPT-4o API pricing as of 2026-09-14, assuming uncached inputs.} In contrast, \framework learns path rewards directly from answer labels without any LLM calls for supervision construction, taking only 2.6 and 3.2 seconds per training question on WebQSP and CWQ (including data processing and reward model training).

\begin{table}[tbp]
\centering
\caption{Efficiency comparison on WebQSP and CWQ. All results are averages per question. \texttt{Train Sup.} denotes training supervision construction; \texttt{\# Tokens} is the sum of input and output tokens.}
\label{tab:efficiency}
\vspace{-2mm}
\resizebox{0.9\textwidth}{!}{
\begin{tabular}{clcccccc}
\toprule
\multirow{2}{*}{\textbf{Stage}}
& \multirow{2}{*}{\textbf{Method}}
& \multicolumn{3}{c}{\textbf{WebQSP}}
& \multicolumn{3}{c}{\textbf{CWQ}} \\
\cmidrule(lr){3-5}\cmidrule(lr){6-8}
& & \textbf{Runtime (s)} & \textbf{\# Calls} & \textbf{\# Tokens}
  & \textbf{Runtime (s)} & \textbf{\# Calls} & \textbf{\# Tokens} \\
\midrule

\multirow{5}{*}{Inference}
& GCR
& 3.5 & \uline{2.0} & \uline{686.6}
& 3.6 & \uline{2.0} & \uline{741.3} \\
& SubgraphRAG
& \uline{2.3} & \textbf{1.0} & 2,699.9
& \uline{2.8} & \textbf{1.0} & 2,607.7 \\
& DP
& 4.4 & 2.5 & 2,799.5
& 6.6 & 3.1 & 3,985.7 \\
\cmidrule{2-8}
& \framework
& \textbf{1.8} & \uline{2.0} & \textbf{382.0}
& \textbf{2.1} & 2.6 & \textbf{593.8} \\

\midrule
\multirow{2}{*}{Train Sup.}
& RAPL
& 4.7 & 1.0 & 2,122.4
& 5.2 & 1.0 & 3,359.7 \\
& \framework
& \textbf{2.6} & \textbf{0} & \textbf{0}
& \textbf{3.2} & \textbf{0} & \textbf{0} \\
\bottomrule
\end{tabular}
}
\end{table}

\subsection{Ablation Study (RQ2--RQ4)}\label{sec:ablation}

We conduct ablation studies with several variants on the test sets (GPT-4o as the answer LLM) to assess the path reward design and the two-stage training strategy, and whether the learned supervision transfers to other retrieval models. Table~\ref{tab:ablation} summarizes the results, with Hits@1 and the implementation details of all variants provided in Appendix~\ref{app:full_ablation}.

\begin{wraptable}{r}{0.49\textwidth}
\vspace{-1.2em}
\centering
\caption{Ablation and transferability analysis on WebQSP and CWQ.}
\label{tab:ablation}
\vspace{-0.6em}
\resizebox{\linewidth}{!}{
\begin{tabular}{lcccc}
\toprule
\multirow[b]{2}{*}{\textbf{Variant}}
& \multicolumn{2}{c}{\textbf{WebQSP}}
& \multicolumn{2}{c}{\textbf{CWQ}} \\
\cmidrule(lr){2-3} \cmidrule(lr){4-5}
& \textbf{F1} & \textbf{Hit}
& \textbf{F1} & \textbf{Hit} \\
\midrule
\framework
& \textbf{81.9} & \uline{91.0}
& \textbf{62.8} & \textbf{71.1} \\

\midrule
\rowcolor{gray!12}
\multicolumn{5}{l}{\textbf{\textit{Path Reward Design}}} \\
\midrule
Pointwise BCE
& 75.9 & 89.4 & 57.5 & 65.7 \\
Pairwise ranking
& 76.1 & 89.3 & 57.9 & 65.5 \\
w/o question
& 62.6 & 80.7 & 46.8 & 58.4 \\

\midrule
\rowcolor{gray!12}
\multicolumn{5}{l}{\textbf{\textit{Reward Transferability}}} \\
\midrule
RoG
& 67.0 & 81.9 & 53.6 & 59.2 \\
\quad + \framework Reward Sup.
& 72.4 & 82.8 & 57.7 & 64.9 \\
SubgraphRAG
& 76.1 & 89.9 & 59.2 & 67.3 \\
\quad + \framework Reward Sup.
& 78.2 & 89.5 & 60.5 & 68.6 \\

\midrule
\rowcolor{gray!12}
\multicolumn{5}{l}{\textbf{\textit{Generator Optimization}}} \\
\midrule
Distillation only
& 80.8 & 90.9
& \uline{61.5} & \uline{70.0} \\
GRPO only
& 44.9 & 61.7 & 37.2 & 54.6 \\
Distillation + GRPO-Ans
& 80.6 & \textbf{91.2} & 61.0 & 69.7 \\
Distillation + GRPO-F1
& \uline{80.9} & 90.6 & 60.4 & 67.9 \\
SFT + GRPO-Ans
& 77.8 & 90.1 & 59.5 & 65.8 \\
\bottomrule
\end{tabular}
}
\vspace{-1.55em} 
\end{wraptable}

\textbf{Reward Model Design (RQ2).} 
We compare \framework with three reward model variants sharing its architecture and training setup, the first two of which treat every answer-reaching path as positive: (1) \textit{Pointwise BCE}, which adapts SubgraphRAG's pointwise classification objective to candidate paths~\citep{li_subgraphrag_2025}; (2) \textit{Pairwise ranking}, which uses a margin loss to rank answer-reaching paths above non-answer-reaching paths~\citep{lin_rje_2025}; and (3) \textit{w/o question}, which removes question conditioning while retaining the asymmetric objective.

Among these alternatives, \textit{Pairwise ranking} performs best, but still consistently underperforms \framework on both WebQSP and CWQ. Removing the question from reward modeling leads to a much larger drop, indicating that the learned reward captures question-dependent path utility rather than a question-agnostic preference. The consistent gains over both conventional objectives further suggest that the asymmetric objective yields more reliable path supervision and, in turn, improved downstream KGQA performance.

\textbf{Reward Transferability (RQ3).} The learned supervision also transfers to other retrieval models. Replacing RoG's uniform shortest-path targets with the highest-scoring answer-reaching path and reweighting SubgraphRAG's path supervision by the learned reward, without changing either architecture, improves F1 by 5.4 and 2.1 points on WebQSP and by 4.1 and 1.3 points on CWQ. Consistent gains on both an LLM-based path generator and a lightweight triple retriever indicate that the learned reward provides reusable intermediate supervision across retrieval architectures.

\textbf{Reward-Guided Path Generation (RQ4).} We consider five variants to characterize the roles of reward distillation and on-policy optimization: (1) \textit{Distillation only}, which removes on-policy optimization; (2) \textit{GRPO only}, which applies GRPO with the learned reward to the base model, without distillation; (3) \textit{Distillation + GRPO-Ans}, which replaces the learned GRPO reward with a binary answer-reaching reward; (4) \textit{Distillation + GRPO-F1}, which uses answer-set F1 as the GRPO reward; and (5) \textit{SFT + GRPO-Ans}, which replaces distillation with uniform SFT over answer-reaching paths and uses the binary answer reward in GRPO.

Distillation alone provides a strong initialization, whereas applying GRPO with the learned reward directly to the base model, without distillation, leads to a substantial performance drop. Starting from the distilled policy, on-policy optimization with the learned reward further improves F1 by 1.1 and 1.3 points on WebQSP and CWQ. The reward used in this stage is decisive: replacing it with the binary answer reward or answer-set F1 lowers F1 by 1.3 and 1.8 points and by 1.0 and 2.4 points relative to the full model, and neither variant improves over distillation alone. Removing the learned reward from both stages, which reduces to the standard SFT-then-GRPO recipe, causes the largest decline among the distilled variants, 4.1 and 3.3 points below the full model, confirming that the learned reward is essential to both initialization and policy refinement.

\subsection{Analysis of Learned Path Rewards (RQ5)}\label{sec:reward_analysis}

We further analyze the learned path rewards from two angles: how accurately they identify the reference path among answer-reaching candidates, and how this accuracy holds up as the number of such candidates grows. Both analyses use training questions with more than one answer-reaching candidate path, the setting in which answer-derived supervision is ambiguous. To limit the cost of LLM-based scoring, we randomly sample 200 such questions per dataset and report results averaged over five runs, with results on the development sets provided in Appendix~\ref{app:additional_path_ranking}.

\begin{wraptable}{r}{0.5\textwidth}
\vspace{-1em}
\centering
\caption{SPARQL path Hits@1 and MRR (\%).}
\vspace{-2mm}
\label{tab:reward_quality}
\resizebox{\linewidth}{!}{
\begin{tabular}{lcccc}
\toprule
\multirow[b]{2}{*}{\textbf{Scoring Method}}
& \multicolumn{2}{c}{\textbf{WebQSP}}
& \multicolumn{2}{c}{\textbf{CWQ}} \\
\cmidrule(lr){2-3} \cmidrule(lr){4-5}
& \textbf{Hits@1} & \textbf{MRR}
& \textbf{Hits@1} & \textbf{MRR} \\
\midrule
Cosine sim.
& 56.9 & 70.2 & 29.3 & 44.3 \\
Answer F1
& 71.3 & 81.6 & 22.2 & 38.9 \\
Pointwise BCE
& 55.7 & 70.4 & 25.7 & 37.2 \\
Pairwise ranking
& 72.8 & 81.4 & 30.0 & 41.5 \\
GPT-4o
& \uline{88.6} & \uline{92.3} & \uline{56.5} & \uline{68.1} \\
\midrule
\framework Reward
& \textbf{90.2} & \textbf{93.1}
& \textbf{71.8} & \textbf{78.7} \\
\bottomrule
\end{tabular}
}
\end{wraptable}

\textbf{Training Supervision Quality.} Table~\ref{tab:reward_quality} ranks the answer-reaching candidate paths of each question with different scorers and reports SPARQL path Hits@1 and MRR against the reference paths. We compare two heuristic scorers, cosine similarity between question and path embeddings and Answer F1, which scores each path by the F1 of its retrieved entities against the ground-truth answers. We also include the two conventional learning objectives from Section~\ref{sec:ablation} and GPT-4o as an LLM-based scorer. \framework outperforms all of them on both datasets. Notably, Answer F1, despite using the ground-truth answers directly, trails \framework by 18.9 and 49.6 Hits@1 points, showing that answer overlap alone cannot identify the reference path and explaining why answer-derived labels are noisy. GPT-4o comes within 1.6 Hits@1 points of \framework on WebQSP, but on CWQ, with more hops and larger candidate sets, \framework leads by 15.3 Hits@1 and 10.6 MRR points. These results support the effectiveness of the learned reward as intermediate supervision without LLM feedback.

\begin{wrapfigure}[13]{r}{0.44\textwidth} 
\vspace{-1.5em} 
\centering
\includegraphics[width=\linewidth]{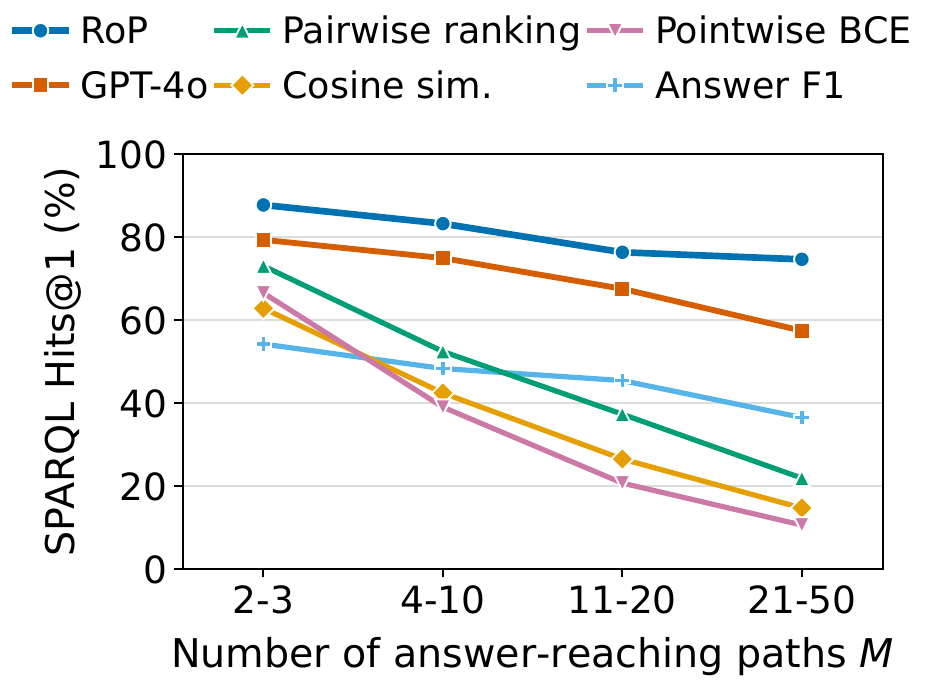}
\caption{SPARQL path Hits@1 under varying levels of supervision ambiguity.} \label{fig:reward_ambiguity} 
\vspace{-3.2em} 
\end{wrapfigure}
\textbf{Effect of Supervision Ambiguity.} Figure~\ref{fig:reward_ambiguity} groups the pooled WebQSP and CWQ samples by the number of answer-reaching candidate paths $M$, a proxy for supervision ambiguity, with a larger $M$ indicating more paths consistent with the answer labels. Hits@1 decreases for all scorers as $M$ grows, but \framework degrades most gradually, and its lead over every baseline, including the GPT-4o scorer, widens accordingly. The widening gap further highlights the advantage of \framework in distinguishing among answer-reaching paths as supervision ambiguity increases.

Supplementary results, case studies, and error analyses are provided in Appendices~\ref{app:sup_result}--\ref{app:error_analysis}.

\section{Conclusion}\label{sec:conclusion}
We proposed \framework, a framework for learning path rewards from answer labels to train an LLM-based relation path generator for KGQA. By combining asymmetric path reward learning with reward distillation and on-policy optimization, \framework provides more reliable intermediate supervision for path generation without costly LLM-refined supervision construction. Experiments on multiple KGQA datasets demonstrate strong KGQA and retrieval accuracy, while further analyses show that the learned rewards remain effective under ambiguous answer-derived supervision and transfer across KG retrieval models. These results highlight the effectiveness of learned path rewards as reliable and scalable intermediate supervision for KG-grounded reasoning.

\subsection*{AI use statement}

We used generative AI tools to assist in implementing methods, specifically to write and debug parts of the experimental code, and did not use them for any other task requiring disclosure, including developing the method, formulating claims, designing experiments, or interpreting results. Additionally, we used generative AI tools to edit the paper for readability. All AI-generated code was verified and tested by the authors, and all AI-edited text was checked against the methods and results it describes. We take responsibility for the final content of this work, including text, claims, or artifacts produced with the aid of generative AI.

\subsection*{Ethics statement}

This work uses existing KGQA benchmarks and knowledge graph resources and does not involve human participants, user studies, or the collection of new personal data. We do not identify additional ethical concerns specific to the proposed methodology beyond those associated with the underlying data and models.

\subsection*{Reproducibility statement}

Implementation details and hyperparameters are provided in Appendix~\ref{app:implementation}, prompt templates in Appendix~\ref{app:prompt}, dataset statistics in Appendix~\ref{app:datasets}, and baseline descriptions and protocols in Appendix~\ref{app:baselines}. Anonymous source code and experimental configurations are available at \url{https://anonymous.4open.science/r/Reward-on-Path-0CB2}.


\bibliography{references}

@article{shao_grpo_2024,
  title={{DeepSeekMath}: Pushing the Limits of Mathematical Reasoning in Open Language Models},
  author={Shao, Zhihong and Wang, Peiyi and Zhu, Qihao and Xu, Runxin and Song, Junxiao and Bi, Xiao and Zhang, Haowei and Zhang, Mingchuan and Li, Y. K. and Wu, Yang and others},
  journal={arXiv preprint arXiv:2402.03300},
  year={2024}
}

@article{yang_qwen3_2025,
  title={Qwen3 Technical Report},
  author={Yang, An and Li, Anfeng and Yang, Baosong and Zhang, Beichen and Hui, Binyuan and Zheng, Bo and Yu, Bowen and Gao, Chang and Huang, Chengen and Lv, Chenxu and others},
  journal={arXiv preprint arXiv:2505.09388},
  year={2025}
}

@article{vrandecic_wikidata_2014,
  author = {Vrande{\v c}i{\'c}, Denny and Kr{\"o}tzsch, Markus},
  year = {2014},
  title = {Wikidata: A Free Collaborative Knowledgebase},
  journal = {Communications of the ACM},
  volume = {57},
  pages = {78--85}
}

@inproceedings{bollacker_freebase_2008,
  author = {Bollacker, Kurt and Evans, Colin and Paritosh, Praveen and Sturge, Tim and Taylor, Jamie},
  year = {2008},
  title = {Freebase: A Collaboratively Created Graph Database for Structuring Human Knowledge},
  booktitle = {{{SIGMOD}}},
  pages = {1247--1250}
}

@inproceedings{auer_dbpedia_2007,
  author = {Auer, Sören and Bizer, Christian and Kobilarov, Georgi and Lehmann, Jens and Cyganiak, Richard and Ives, Zachary},
  title = {{{DBpedia}}: {{A Nucleus}} for a {{Web}} of {{Open Data}}},
  booktitle = {The {{Semantic Web}}},
  pages = {722--735},
  year = {2007},
}

@article{pan_unifyingSurvey_2024,
  author = {Pan, Shirui and Luo, Linhao and Wang, Yufei and Chen, Chen and Wang, Jiapu and Wu, Xindong},
  title = {Unifying Large Language Models and Knowledge Graphs: A Roadmap},
    volume={36},
  number={7},
  pages={3580--3599},
  year = {2024},
  journal = {IEEE Transactions on Knowledge and Data Engineering}
}

@inproceedings{li_subgraphrag_2025,
  author = {Li, Mufei and Miao, Siqi and Li, Pan},
  year = {2025},
  title = {Simple is Effective: The Roles of Graphs and Large Language Models in Knowledge-Graph-Based Retrieval-Augmented Generation},
  booktitle = {ICLR}
}

@inproceedings{mavromatis_gnnrag_2025,
  title={{GNN-RAG}: Graph Neural Retrieval for Efficient Large Language Model Reasoning on Knowledge Graphs},
  author={Mavromatis, Costas and Karypis, George},
  booktitle={Findings of ACL},
  year={2025},
  pages={16682--16699}
}

@inproceedings{luo_gcr_2025,
  title={Graph-constrained Reasoning: Faithful Reasoning on Knowledge Graphs with Large Language Models},
  author={Luo, Linhao and Zhao, Zicheng and Haffari, Gholamreza and Li, Yuan-Fang and Gong, Chen and Pan, Shirui},
  booktitle={ICML},
  pages={41540--41565},
  year={2025}
}

@inproceedings{luo_rog_2024,
  title={Reasoning on graphs: Faithful and interpretable large language model reasoning},
  author={Luo, Linhao and Li, Yuan-Fang and Haffari, Gholamreza and Pan, Shirui},
  booktitle={ICLR},
  year={2024}
}

@inproceedings{sun_tog_2024,
  title={{Think-on-Graph}: Deep and responsible reasoning of large language model on knowledge graph},
  author={Sun, Jiashuo and Xu, Chengjin and Tang, Lumingyuan and Wang, Saizhuo and Lin, Chen and Gong, Yeyun and Ni, Lionel M and Shum, Heung-Yeung and Guo, Jian},
  booktitle={ICLR},
  year={2024}
}

@inproceedings{lin_rje_2025,
    title = "{RJE}: A Retrieval-Judgment-Exploration Framework for Efficient Knowledge Graph Question Answering with {LLM}s",
    author = "Lin, Can  and
      Jiang, Zhengwang  and
      Zheng, Ling  and
      Zhao, Qi  and
      Zhang, Yuhang  and
      Song, Qi  and
      Zhou, Wangqiu",
    booktitle = "EMNLP",
    year = "2025",
    pages = "17277--17294",
}

@inproceedings{um_rporag_2026,
  title={{RPO-RAG}: Aligning Small {LLMs} with Relation-aware Preference Optimization for Knowledge Graph Question Answering},
  author={Um, Kaehyun and Yeom, KyuHwan and Yang, Haerim and Choi, Minyoung and Yang, Hyeongjun and Lee, Kyong-Ho},
  booktitle={WWW},
  pages={2513--2523},
  year={2026}
}

@article{yao_rapl_2025,
  title={Learning Efficient and Generalizable Graph Retriever for Knowledge-Graph Question Answering},
  author={Yao, Tianjun and Li, Haoxuan and Shen, Zhiqiang and Li, Pan and Liu, Tongliang and Zhang, Kun},
  journal={arXiv preprint arXiv:2506.09645},
  year={2025}
}

@inproceedings{wang_damr_2026,
  title={{DAMR}: Efficient and Adaptive Context-Aware Knowledge Graph Question Answering with {LLM}-Guided {MCTS}},
  author={Wang, Yingxu and Fan, Shiqi and Wang, Mengzhu and Gao, Siyang and Wang, Chao and Yin, Nan},
  booktitle={ICLR},
  year={2026}
}

@inproceedings{lin_kgqarl_2018,
    title = "Multi-Hop Knowledge Graph Reasoning with Reward Shaping",
    author = "Lin, Xi Victoria  and
      Socher, Richard  and
      Xiong, Caiming",
    booktitle = "EMNLP",
    year = "2018",
    pages = "3243--3253",
}

@article{lu_double_2026,
  title={{Double-Calibration}: Towards Reliable {LLMs} via Calibrating Knowledge and Reasoning Confidence},
  author={Lu, Yuyin and Liang, Ziran and Rao, Yanghui and Fan, Wenqi and Wang, Fu Lee and Li, Qing},
  journal={arXiv preprint arXiv:2601.11956},
  year={2026}
}

@inproceedings{luo_kbqao1_2025,
  author={Luo, Haoran and E, Haihong and Guo, Yikai and Lin, Qika and Wu, Xiaobao and Mu, Xinyu and Liu, Wenhao and Song, Meina and Zhu, Yifan and Luu, Anh Tuan},
  year={2025},
  title={{KBQA-o1}: Agentic Knowledge Base Question Answering with {Monte Carlo Tree Search}},
  booktitle={ICML},
  pages={41177--41199}
}

@inproceedings{ma_dp_2025,
  title={Deliberation on Priors: Trustworthy Reasoning of Large Language Models on Knowledge Graphs}, 
  author={Jie Ma and Ning Qu and Zhitao Gao and Rui Xing and Jun Liu and Hongbin Pei and Jiang Xie and Linyun Song and Pinghui Wang and Jing Tao and Zhou Su},
  booktitle={NeurIPS},
  year={2025}
}

@inproceedings{chen_pog_2024,
  title={{Plan-on-Graph}: Self-Correcting Adaptive Planning of Large Language Model on Knowledge Graphs},
  author={Chen, Liyi and Tong, Panrong and Jin, Zhongming and Sun, Ying and Ye, Jieping and Xiong, Hui},
  booktitle={NeurIPS},
  pages={37665--37691},
  year={2024}
}

@inproceedings{li_kbBinder_2023,
	title = {Few-shot In-context Learning for Knowledge Base Question Answering},
	author = {Li, Tianle and Ma, Xueguang and Zhuang, Alex and Gu, Yu and Su, Yu and Chen, Wenhu},
    booktitle = {ACL},
	year = {2023},
    pages = {6966--6980}
}

@inproceedings{li_flexkbqa_2024,
	title = {{FlexKBQA}: {A} Flexible {LLM}-Powered Framework for Few-Shot Knowledge Base Question Answering},
	author = {Li, Zhenyu and Fan, Sunqi and Gu, Yu and Li, Xiuxing and Duan, Zhichao and Dong, Bowen and Liu, Ning and Wang, Jianyong},
    booktitle = {AAAI},
	year = {2024},
    pages = {18608--18616}
}

@article{zou_reg_2025,
  title={Weak-to-strong {GraphRAG}: Aligning weak retrievers with large language models for graph-based retrieval augmented generation},
  author={Zou, Deyu and Chen, Yongqiang and Li, Mufei and Miao, Siqi and Liu, Chenxi and Han, Bo and Cheng, James and Li, Pan},
  journal={arXiv preprint arXiv:2506.22518},
  year={2025}
}

@article{wang_kg-hopper_2026,
  title={{KG-Hopper}: Empowering Compact Open {LLMs} with Knowledge Graph Reasoning via Reinforcement Learning},
  author={Wang, Shuai and Yu, Yinan},
  journal={arXiv preprint arXiv:2603.21440},
  year={2026}
}

@inproceedings{zhang_metaqa_2018,
  title={Variational Reasoning for Question Answering with Knowledge Graph},
  author={Zhang, Yuyu and Dai, Hanjun and Kozareva, Zornitsa and Smola, Alexander J. and Song, Le},
  booktitle={AAAI},
  pages={6069--6076},
  year={2018}
}

@inproceedings{sui_fidelis_2025,
  author={Sui, Yuan and He, Yufei and Liu, Nian and He, Xiaoxin and Wang, Kun and Hooi, Bryan},
  title={{FiDeLiS}: Faithful Reasoning in Large Language Models for Knowledge Graph Question Answering},
  year={2025},
  booktitle={Findings of ACL},
  pages={8315--8330}
}

@inproceedings{ma_debate_2025,
  title={Debate on Graph: A Flexible and Reliable Reasoning Framework for Large Language Models},
  author={Ma, Jie and Gao, Zhitao and Chai, Qi and Sun, Wangchun and Wang, Pinghui and Pei, Hongbin and Tao, Jing and Song, Lingyun and Liu, Jun and Zhang, Chen and others},
  booktitle={AAAI},
  pages={24768--24776},
  year={2025}
}

@article{lan_complexkbqa_2023,
	title = {Complex Knowledge Base Question Answering: {A} Survey},
	volume = {35},
	number = {11},
	journal = {IEEE Transactions on Knowledge and Data Engineering},
	author = {Lan, Yunshi and He, Gaole and Jiang, Jinhao and Jiang, Jing and Zhao, Wayne Xin and Wen, Ji-Rong},
	year = {2023},
	pages = {11196--11215}
}

@inproceedings{gu_beyond_2021,
	title = {Beyond {I}.{I}.{D}.: {Three} Levels of Generalization for Question Answering on Knowledge Bases},
	shorttitle = {Beyond {I}.{I}.{D}.},
	booktitle = {WWW},
	author = {Gu, Yu and Kase, Sue and Vanni, Michelle and Sadler, Brian and Liang, Percy and Yan, Xifeng and Su, Yu},
	year = {2021},
	pages = {3477--3488}
}

@inproceedings{yih_webqsp_2016,
  author = {Yih, Wen-tau and Richardson, Matthew and Meek, Chris and Chang, Ming-Wei and Suh, Jina},
  year = {2016},
  title = {The Value of Semantic Parse Labeling for Knowledge Base Question Answering},
  booktitle = {{{ACL}}},
  pages = {201--206}
}

@inproceedings{talmor_cwq_2018,
  author = {Talmor, Alon and Berant, Jonathan},
  year = {2018},
  title = {The Web as a Knowledge-Base for Answering Complex Questions},
  booktitle = {{{NAACL}}},
  pages = {641--651}
}

@article{li_gte_2023,
  title={Towards general text embeddings with multi-stage contrastive learning},
  author={Li, Zehan and Zhang, Xin and Zhang, Yanzhao and Long, Dingkun and Xie, Pengjun and Zhang, Meishan},
  journal={arXiv preprint arXiv:2308.03281},
  year={2023}
}

@inproceedings{hu_lora_2021,
	title = {{LoRA}: {Low}-Rank Adaptation of Large Language Models},
	author = {Hu, Edward J. and Shen, Yelong and Wallis, Phillip and Allen-Zhu, Zeyuan and Li, Yuanzhi and Wang, Shean and Wang, Lu and Chen, Weizhu},
	year = {2022},
	booktitle = {ICLR}
}
\bibliographystyle{iclr2027_conference}

\appendix
\section{Relation Path Example}\label{app:path_example}

We provide an illustrative example on how a relation path is grounded over a KG.
Consider a relation path
$z = \langle \text{sibling} \rightarrow \text{nationality} \rangle$.
Given a starting entity $e=\text{Elon Musk}$, the reachable entity set
$\mathcal{E}_z(e)$ contains all entities that can be reached by following the relations in $z$ from $e$.
Traversing the path $z$ from the entity \text{Elon Musk} can ground a KG reasoning path:
\[
\text{Elon Musk}
\xrightarrow{sibling}
\text{Kimbal Musk}
\xrightarrow{nationality}
\{\text{USA, Canada, South Africa}\}.
\]
Thus,
\[
\mathcal{E}_z(\text{Elon Musk})
=
\{\text{USA, Canada, South Africa}\}.
\]

\section{Prompt Template}\label{app:prompt}

In this section, we illustrate all prompt templates used in this work.

\textbf{Relation Path Generation Prompt.} As shown in Figure~\ref{fig:path_gen_prompt}, the prompt template for relation path generation takes a question and its question entity as input, and instructs the LLM to generate a relation path consisting of a sequence of relations, i.e.,  $z=\langle r_1,r_2,\ldots,r_l\rangle$. The same template is used in both training stages and at inference.

\textbf{KG-grounded Answer Reasoning Prompt.}
 As shown in Figure~\ref{fig:reasoning_prompt}, the answer reasoning prompt takes the question and the evidence context $\mathcal{C}_q$, i.e., the grounded paths of all question entities, each listed with its question entity, relation path, and retrieved end entities.

\begin{figure}[h]
    \centering
    \begin{tcolorbox}[
        colback=gray!5,
        colframe=black!70,
        colbacktitle=black!70,
        coltitle=white,
        title=Relation Path Generation Prompt,
        fonttitle=\bfseries,
        arc=2pt,
        boxrule=0.7pt,
        left=10pt,
        right=10pt,
        top=6pt,
        bottom=6pt,
        titlerule=0pt,
        width=0.95\linewidth
    ]
    \textbf{Instruction:} Reasoning path is a sequence of relations in the Knowledge Graph that connects the topic entity in the question to answer entities. Given a question, please generate a reasoning path in the Knowledge Graph starting from the topic entity to answer the question.

    \vspace{0.5em}

    \textbf{Input:}

    Question: \texttt{<Question>}

    Topic entity: \texttt{<Topic Entity>}

    \vspace{1em}

    \textbf{Output:}

    \texttt{<PATH>} $r_1 \rightarrow r_2 \rightarrow \cdots \rightarrow r_l$ \texttt{</PATH>}
    \end{tcolorbox}

    \caption{The prompt template for relation path generation.}
    \label{fig:path_gen_prompt}
\end{figure}

\begin{figure}[h]
    \centering
    \begin{tcolorbox}[
        colback=gray!5,
        colframe=black!70,
        colbacktitle=black!70,
        coltitle=white,
        title=KG-Grounded Answer Reasoning,
        fonttitle=\bfseries,
        arc=2pt,
        boxrule=0.7pt,
        left=10pt,
        right=10pt,
        top=6pt,
        bottom=6pt,
        titlerule=0pt,
        width=0.95\linewidth
    ]
    \textbf{Instruction:} You are a helpful and precise assistant for answering questions based on the provided reasoning paths on a knowledge graph. Please return all the possible answers from the entities mentioned in the reasoning paths. Please return each answer at a new line.

    \vspace{0.5em}

    \textbf{Input:}

    \vspace{0.5em}
    
    Reasoning Paths: 

    \vspace{0.5em}
    
    \texttt{[PATH1]}
    
    Topic Entity: \texttt{<Topic Entity>}, 
    
    Relation Path: $r_1 \rightarrow r_2 \rightarrow \cdots \rightarrow r_l$ 
    
    End Entities: \texttt{<End Entity$_1$>} \texttt{<SEP>} \texttt{<End Entity$_2$>} \texttt{<SEP>} $\cdots$

    \vspace{0.5em}
    
    \texttt{[PATH2]}
    
    $\cdots$
    
    \vspace{0.5em}
    
    Question: \texttt{<Question>}

    \vspace{1em}

    \textbf{Output:}
    \vspace{0.5em}

    \texttt{Answer$_1$}
    
    \texttt{Answer$_2$}
    
    $\cdots$
    
    \texttt{Answer$_n$}
    \end{tcolorbox}

    \caption{The prompt template for KG-grounded inductive reasoning.}
    \label{fig:reasoning_prompt}
\end{figure}

\section{Datasets Statistics}\label{app:datasets}

Table~\ref{tab:dataset_statistics} reports the overall scale of each dataset, including the associated KG, numbers of entities, relations, triples, training and test questions, and the maximum reasoning hop. Table~\ref{tab:answer_number_statistics} provides the distribution of answer set sizes for each dataset, which helps characterize the answer sparsity and complexity of these datasets.

\begin{table}[htbp]
\centering
\caption{Statistics of the datasets.}
\label{tab:dataset_statistics}
\resizebox{\textwidth}{!}{
\begin{tabular}{llrrrrrrr}
\toprule
\textbf{Dataset} & \textbf{KG} & \textbf{Entities} & \textbf{Relations} & \textbf{Triples} & \textbf{Train} & \textbf{Test} & \textbf{Max Hop} & \textbf{License} \\
\midrule
WebQSP & Freebase & 2,566,291 & 7,058 & 8,309,105 & 2,826 & 1,628 & 2 & CC BY 4.0\\
CWQ & Freebase & 2,566,291 & 7,058 & 8,309,105 & 27,639 & 3,531 & 4 & -\\
MetaQA & Wiki-Movie & 43,234 & 9 & 133,592 & 329,282 & 30,903 & 3 & -\\
\bottomrule
\end{tabular}
}
\end{table}

\begin{table}[htbp]
\centering
\small
\caption{Statistics of number of answers on WebQSP and CWQ.}
\label{tab:answer_number_statistics}
\begin{tabular}{lcccc}
\toprule
\textbf{Dataset} 
& \textbf{\#Ans = 1} 
& $\mathbf{2 \leq \#}\textbf{Ans}\mathbf{\leq 4}$ 
& $\mathbf{5 \leq \#}\textbf{Ans}\mathbf{\leq 9}$ 
& $\mathbf{\#}\textbf{Ans}\mathbf{\geq 10}$ \\
\midrule
WebQSP & 51.2\% & 27.4\% & 8.3\% & 12.1\% \\
CWQ & 70.6\% & 19.4\% & 6\% & 4\% \\
MetaQA & 37.9\% & 29.6\% & 12.6\% & 19.9\% \\
\bottomrule
\end{tabular}
\end{table}

\section{Baselines}\label{app:baselines}

We group the baselines according to how they obtain supervision for KG reasoning: (1) \textit{prompting-based methods}, which perform KG reasoning through LLM prompting without task-specific training; (2) \textit{training with answer-derived supervision}, which derives training signals from answer labels; and (3) \textit{training with LLM-refined supervision}, which uses LLMs to construct, refine, or evaluate intermediate reasoning signals. We describe each baseline below.

\textbf{Prompting-based Methods.}
These methods rely on prompted LLMs to explore or reason over the KG without task-specific model training.

\begin{itemize}
    \item ToG~\citep{sun_tog_2024} treats the LLM as an agent that performs beam search over the KG, iteratively expanding relations and entities until sufficient evidence is collected for answering.

    \item DoG~\citep{ma_debate_2025} employs multiple prompted LLM roles to debate over candidate subgraphs, with an answer-trying module determining whether to stop or continue graph exploration.

    \item FiDeLiS~\citep{sui_fidelis_2025} combines training-free path retrieval with deductive verification, verifying each reasoning step against the KG before proceeding or producing the final answer.
\end{itemize}

\textbf{Training with Answer-derived Supervision.}
These methods derive supervision from ground-truth answer labels alone.

\begin{itemize}
    \item RoG~\citep{luo_rog_2024} adopts a planning--retrieval--reasoning pipeline in which an LLM generates relation paths, retrieves matching KG reasoning paths, and reasons over the retrieved evidence for answer generation.

    \item GNN-RAG~\citep{mavromatis_gnnrag_2025} uses a GNN to score answer candidates over a question-specific subgraph and verbalizes shortest paths from question entities to predicted candidates for LLM-based answer reasoning.

    \item GCR~\citep{luo_gcr_2025} incorporates KG structure into LLM decoding through a KG-Trie, using a KG-specialized LLM to generate KG-grounded reasoning paths and a general LLM to aggregate them for answer prediction.

    \item SubgraphRAG~\citep{li_subgraphrag_2025} retrieves compact evidence subgraphs by scoring KG triples with a lightweight model enhanced by directional distance encoding, and prompts an LLM to reason over the retrieved evidence.

    \item DP~\citep{ma_dp_2025} distills KG structural priors into an LLM through supervised fine-tuning and preference optimization, and employs reasoning introspection to verify constraints and trigger backtracking when necessary.

    \item RPO-RAG~\citep{um_rporag_2026} ranks answer-reaching paths by their similarity to the question under a pretrained semantic encoder and incorporates the resulting preferences into relation-aware preference optimization.
\end{itemize}

\textbf{Training with LLM-refined Supervision.} These methods use LLMs to refine intermediate supervision or to provide reward signals for training KG reasoning models.

\begin{itemize}
    \item KG-Hopper~\citep{wang_kg-hopper_2026} trains a compact LLM for KG reasoning with reinforcement learning, using LLM-based rewards to evaluate generated reasoning paths during policy optimization.

    \item RAPL~\citep{yao_rapl_2025} trains a lightweight graph retriever with LLM-rationalized path labels and triple-level message passing to retrieve compact KG evidence for downstream reasoning.

    \item ReG~\citep{zou_reg_2025} refines weak supervision with LLM feedback and reorganizes retrieved graph evidence into logic-preserving chains for downstream LLM reasoning.
\end{itemize}

\section{Implementation Details}\label{app:implementation}

All experiments are conducted on a machine with two NVIDIA H100 GPUs and 320~GB RAM. Unless otherwise specified, we use the preprocessed Freebase subgraphs described in Section~\ref{sec:experiment}.

\subsection{Path Reward Learning}

We construct candidate relation paths by breadth-first search from each question entity, with the maximum path length set to $L=2$ for WebQSP and $L=3$ for CWQ. We retain all candidate paths that reach at least one ground-truth answer. To control the candidate set size and reduce the computational cost of reward-model training, we randomly sample 50 non-answer-reaching paths per training question on WebQSP and 200 on CWQ.

For the path reward model, we use gte-large-en-v1.5~\citep{li_gte_2023} as the frozen text encoder for both questions and relations. The resulting embeddings are processed by a two-layer Transformer encoder with a hidden dimension of 128 and an MLP scoring head. The text encoder remains frozen, while the Transformer encoder, positional embeddings, and MLP are trainable, resulting in 0.5M trainable parameters.

We optimize the asymmetric reward-learning objective with AdamW using a learning rate of $4\times10^{-4}$, weight decay of $0.01$, and gradient clipping at $1.0$. The batch size is 4 question--entity pairs per optimization step. We linearly warm up the learning rate for the first 10\% of training steps and then apply cosine decay to zero. The reward model is trained for 500 epochs on WebQSP and 200 epochs on CWQ.

\subsection{Reward-Guided Relation Path Generation}

We use LLaMA3.1-8B-Instruct as the relation path generator and optimize only its LoRA parameters~\citep{hu_lora_2021}, while keeping the base LLM weights frozen. The LoRA rank and scaling factor are set to 16 and 32, respectively. The learned path reward model remains frozen throughout both stages of generator training.

\textbf{Path Reward Distillation.}
For each training instance, we retain all answer-reaching candidate paths and construct the target distribution by applying a softmax over their learned reward scores. The generator is optimized using the complete autoregressive likelihood of each relation path, without length normalization or truncating the candidate set by reward rank.

We train the generator for 10 epochs with AdamW using a learning rate of $5\times10^{-5}$, weight decay of $0.01$, and an effective batch size of 16. The learning rate is kept constant without warmup.

\textbf{On-Policy Reward Optimization.}
Starting from the distilled generator, we further optimize the policy with GRPO. For each question--entity pair, we sample $G=8$ relation paths using rollout temperatures
$\{1.0,1.0,1.15,1.15,1.3,1.3,1.45,1.45\}$.
Policy probabilities used for computing the importance ratios are evaluated at temperature 1, without additional importance-sampling correction for the rollout temperatures. 

Rewards are normalized within each rollout group with $\epsilon_{\mathrm{std}}=10^{-6}$. We use a PPO clipping coefficient of $\varepsilon=0.2$ and perform one policy-update epoch per rollout batch. The loss is averaged over all generated tokens in the batch, without additional KL-divergence or entropy regularization. We optimize the LoRA parameters with AdamW using a learning rate of $1\times10^{-5}$, weight decay of $0.01$, and gradient clipping at $1.0$; the learning rate is kept constant without warmup. Each update uses 16 prompts with 8 rollouts per prompt, corresponding to 128 generated sequences. We train for approximately 3 epochs over the prompt pool, reporting the checkpoints at step 540 on WebQSP and step 7,893 on CWQ.

\textbf{KG-Grounded Answer Reasoning.} At inference time, we generate the top-$K$ relation paths with beam search using $K=5$. Each generated path is grounded from the corresponding question entity in the KG, and the resulting evidence is provided to the answer LLM through zero-shot prompting with decoding temperature 0.

\section{Supplementary Results}\label{app:sup_result}

This section provides additional experimental results that complement the main evaluation, including run-to-run stability, robustness across model backbones, performance across reasoning hops, additional path-ranking analyses, and extended ablation results.

\subsection{Standard Deviation of KGQA Performance}\label{app:stand_deviation}

\begin{table}[h]
\centering
\small
\caption{Stability analysis of \framework{} with answer LLMs. Results are reported as mean values with standard deviations shown as subscripts.}
\label{tab:deviation}

\begin{tabular}{lcccccc}
\toprule
\multirow{2}{*}{\textbf{LLM}} 
& \multicolumn{3}{c}{\textbf{WebQSP}} 
& \multicolumn{3}{c}{\textbf{CWQ}} \\
\cmidrule(lr){2-4} \cmidrule(lr){5-7}
& \textbf{F1} & \textbf{Hit} & \textbf{Hits@1}
& \textbf{F1} & \textbf{Hit} & \textbf{Hits@1} \\
\midrule
GPT-4o 
& $81.9_{\text{\scriptsize $\pm 0.6$}}$ 
& $91.0_{\text{\scriptsize $\pm 0.5$}}$ 
& $86.0_{\text{\scriptsize $\pm 0.7$}}$
& $62.8_{\text{\scriptsize $\pm 0.5$}}$ 
& $71.1_{\text{\scriptsize $\pm 0.4$}}$ 
& $65.3_{\text{\scriptsize $\pm 0.6$}}$ \\

GPT-4.1
& $82.2_{\text{\scriptsize $\pm 0.6$}}$
& $91.2_{\text{\scriptsize $\pm 0.5$}}$
& $85.2_{\text{\scriptsize $\pm 0.7$}}$
& $62.7_{\text{\scriptsize $\pm 0.5$}}$
& $73.2_{\text{\scriptsize $\pm 0.5$}}$
& $64.3_{\text{\scriptsize $\pm 0.6$}}$ \\
\bottomrule
\end{tabular}
\vspace{-2mm}
\end{table}

Following prior work~\citep{ma_dp_2025}, we conduct three independent runs and report the mean and standard deviation in Table~\ref{tab:deviation}.

\subsection{Performance with Different Backbones}\label{app:backbone}

\begin{table}[h]
\centering
\small
\caption{KGQA performance (\%) of \framework with different text encoders, relation path generators, and answer generators.}
\label{tab:backbone}
\begin{tabular}{lcccccc}
\toprule
\multirow{2}{*}{\textbf{Configuration}}
& \multicolumn{3}{c}{\textbf{WebQSP}}
& \multicolumn{3}{c}{\textbf{CWQ}} \\
\cmidrule(lr){2-4} \cmidrule(lr){5-7}
& \textbf{F1} & \textbf{Hit} & \textbf{Hits@1}
& \textbf{F1} & \textbf{Hit} & \textbf{Hits@1} \\
\midrule
\framework{} (default)
& 81.9 & 91.0 & \textbf{86.0}
& \uline{62.8} & \uline{71.1} & \textbf{65.3} \\
\midrule
\rowcolor{gray!12}
\multicolumn{7}{l}{\textbf{\textit{Text Encoder}}} \\
\midrule
all-MiniLM-L6-v2
& 81.7 & 90.3 & \uline{85.5}
& \textbf{63.2} & 70.6 & \uline{65.1} \\
\midrule
\rowcolor{gray!12}
\multicolumn{7}{l}{\textbf{\textit{Relation Path Generator}}} \\
\midrule
Qwen2.5-7B-Instruct
& \textbf{82.4} & \textbf{91.3} & \textbf{86.0}
& 62.2 & 69.9 & 64.8 \\
Qwen3-8B
& 81.4 & 90.6 & 85.3
& 62.3 & 70.3 & 64.5 \\
\midrule
\rowcolor{gray!12}
\multicolumn{7}{l}{\textbf{\textit{Answer Generator}}} \\
\midrule
GPT-4.1
& \uline{82.2} & \uline{91.2} & 85.2
& 62.7 & \textbf{73.2} & 64.3 \\
\bottomrule
\end{tabular}
\end{table}

To examine the effect of backbone choices, we evaluate \framework with alternative text encoders, relation path generators, and answer generators, varying one component at a time while keeping the remaining configuration unchanged. Specifically, we replace gte-large-en-v1.5 with all-MiniLM-L6-v2, LLaMA3.1-8B-Instruct with Qwen2.5-7B-Instruct or Qwen3-8B~\citep{yang_qwen3_2025}, and GPT-4o with GPT-4.1; the first two alternatives are adopted by recent KGQA baselines~\citep{um_rporag_2026,wang_kg-hopper_2026}.

As shown in Table~\ref{tab:backbone}, \framework maintains comparable performance across all choices. Replacing gte-large-en-v1.5 with all-MiniLM-L6-v2 slightly improves F1 on CWQ, Qwen2.5-7B-Instruct achieves the highest F1 and Hit on WebQSP, and Qwen3-8B performs close to the default configuration. Replacing GPT-4o with GPT-4.1 as the answer generator yields comparable F1 (82.2 and 62.7) and a higher Hit on CWQ. Across all alternatives, F1 differs from the default configuration by at most 0.6 points on either dataset, showing that the gains of \framework do not depend on a particular encoder, generator, or answer LLM.

\subsection{Performance across Reasoning Hops}\label{app:hop_breakdown}

\begin{table}[h]
\centering
\caption{Breakdown of KGQA performance (\%) over \# hops.}
\label{tab:hop_f1_hit}
\resizebox{\textwidth}{!}{
\begin{tabular}{lcccccccccccccc}
\toprule
\multirow{3}{*}{\textbf{Model}} 
& \multicolumn{6}{c}{\textbf{WebQSP-sub}} 
& \multicolumn{8}{c}{\textbf{CWQ-sub}} \\
\cmidrule(lr){2-7} \cmidrule(lr){8-15}
& \multicolumn{2}{c}{\textbf{1}} 
& \multicolumn{2}{c}{\textbf{2}} 
& \multicolumn{2}{c}{\textbf{Overall}} 
& \multicolumn{2}{c}{\textbf{1}} 
& \multicolumn{2}{c}{\textbf{2}} 
& \multicolumn{2}{c}{\textbf{$\geq$3}}
& \multicolumn{2}{c}{\textbf{Overall}} \\
\cmidrule(lr){2-3} 
\cmidrule(lr){4-5}
\cmidrule(lr){6-7}
\cmidrule(lr){8-9} 
\cmidrule(lr){10-11} 
\cmidrule(lr){12-13}
\cmidrule(lr){14-15}
& \textbf{F1} & \textbf{Hit} 
& \textbf{F1} & \textbf{Hit} 
& \textbf{F1} & \textbf{Hit} 
& \textbf{F1} & \textbf{Hit} 
& \textbf{F1} & \textbf{Hit}
& \textbf{F1} & \textbf{Hit} 
& \textbf{F1} & \textbf{Hit} \\
\midrule
SubgraphRAG
& 80.3 & 89.7 
& 74.4 & 83.0 
& 78.3 & 87.5 
& 57.0 & 59.7 
& 63.3 & 68.0 
& 51.6 & 52.0 
& 60.8 & 64.7 \\

GCR 
& 71.7 & 90.1 
& 69.7 & 88.5 
& 71.0 & 89.6 
& 60.6 & 67.3 
& 65.6 & 74.5 
& 47.7 & 51.5 
& 63.1 & 71.1 \\
\midrule
\framework 
& \textbf{84.5} & \textbf{92.4} 
& \textbf{80.7} & \textbf{92.3} 
& \textbf{83.2} & \textbf{92.4} 
& \textbf{66.4} & \textbf{75.5} 
& \textbf{75.0} & \textbf{83.8} 
& \textbf{56.2} & \textbf{64.7} 
& \textbf{71.4} & \textbf{80.3} \\
\bottomrule
\end{tabular}
}
\end{table}

Following SubgraphRAG~\citep{li_subgraphrag_2025}, we further evaluate \framework across questions with different reasoning depths on WebQSP-sub and CWQ-sub, where questions whose answer entities are absent from the KG are excluded, and compare with SubgraphRAG and GCR. As shown in Table~\ref{tab:hop_f1_hit}, \framework achieves the best F1 and Hit in every hop group on both datasets. The gains do not shrink with reasoning depth: on WebQSP-sub, \framework improves F1 over SubgraphRAG by 4.2 and 6.3 points on 1-hop and 2-hop questions, respectively. On CWQ-sub, the F1 improvement over GCR reaches 9.4 points on 2-hop questions, and the Hit improvement over SubgraphRAG reaches 12.7 points on questions requiring three or more hops. The benefit of reward-guided path generation thus extends to questions requiring longer relational reasoning.

\subsection{Ablation Study}\label{app:full_ablation}

\begin{table}[h]
\centering
\small
\caption{Complete ablation results on WebQSP and CWQ. We report F1, Hit, and Hits@1 (\%).}
\label{tab:full_ablation}
\begin{tabular}{lcccccc}
\toprule
\multirow{2}{*}{\textbf{Variant}}
& \multicolumn{3}{c}{\textbf{WebQSP}}
& \multicolumn{3}{c}{\textbf{CWQ}} \\
\cmidrule(lr){2-4} \cmidrule(lr){5-7}
& \textbf{F1} & \textbf{Hit} & \textbf{Hits@1}
& \textbf{F1} & \textbf{Hit} & \textbf{Hits@1} \\
\midrule

\rowcolor{gray!12}
\multicolumn{7}{l}{\textbf{\textit{Reward Model Design}}} \\
\midrule
\framework
& \textbf{81.9} & \uline{91.0} & \textbf{86.0}
& \textbf{62.8} & \textbf{71.1} & \textbf{65.3} \\

Pointwise BCE
& 75.9 & 89.4 & 82.6
& 57.5 & 65.7 & 59.8 \\

Pairwise ranking
& 76.1 & 89.3 & 83.1
& 57.9 & 65.5 & 60.4 \\

w/o question
& 62.6 & 80.7 & 71.9
& 46.8 & 58.4 & 50.9 \\

\midrule
\rowcolor{gray!12}
\multicolumn{7}{l}{\textbf{\textit{Transferability of Learned Reward Supervision}}} \\
\midrule
RoG~\citep{luo_rog_2024}
& 67.0 & 81.9 & 78.3
& 53.6 & 59.2 & 53.7 \\

\quad + \framework Reward Sup.
& 72.4 & 82.8 & 79.5
& 57.7 & 64.9 & 59.1 \\

SubgraphRAG~\citep{li_subgraphrag_2025}
& 76.1 & 89.9 & 84.3
& 59.2 & 67.3 & 60.7 \\

\quad + \framework Reward Sup.
& 78.2 & 89.5 & 84.9
& 60.5 & 68.6 & 62.3 \\

\midrule
\rowcolor{gray!12}
\multicolumn{7}{l}{\textbf{\textit{Reward-Guided Path Generation}}} \\
\midrule
Distillation only
& 80.8 & 90.9 & 84.8
& \uline{61.5} & \uline{70.0} & \uline{63.8} \\

GRPO only
& 44.9 & 61.7 & 52.8
& 37.2 & 54.6 & 42.1 \\

Distillation + GRPO-Ans
& 80.6 & \textbf{91.2} & \uline{85.1}
& 61.0 & 69.7 & 63.6 \\

Distillation + GRPO-F1
& \uline{80.9} & 90.6 & 84.5
& 60.4 & 67.9 & 61.8 \\

SFT + GRPO-Ans
& 77.8 & 90.1 & 82.7
& 59.5 & 65.8 & 61.2 \\

\bottomrule
\end{tabular}
\end{table}

\textbf{Reward Model Baselines.}
We compare our asymmetric reward-learning objective with two existing path-scoring objectives while keeping the scoring architecture $s_\phi(q,z)$ unchanged.

\textit{Pointwise BCE} follows the path-scoring objective used in SubgraphRAG~\citep{li_subgraphrag_2025}. Each candidate path is assigned the answer-derived label
\begin{equation}
y_i
=
\mathbb{I}
\left[
z_i\in\mathcal{Z}_{q,e}^{\mathcal{A}_q}
\right],
\end{equation}
and optimized with binary cross-entropy:
\begin{equation}
\mathcal{L}_{\mathrm{BCE}}^{q,e}
=
-\frac{1}{|\mathcal{Z}_{q,e}|}
\sum_{z_i\in\mathcal{Z}_{q,e}}
\left[
y_i\log\sigma(s_\phi(q,z_i))
+
(1-y_i)\log(1-\sigma(s_\phi(q,z_i)))
\right].
\label{eq:ablation_bce}
\end{equation}

\textit{Pairwise ranking} follows the margin-based path-ranking objective~\citep{lin_rje_2025}. For each question--entity pair, we use all answer-reaching and non-answer-reaching path pairs:
\begin{equation}
\mathcal{P}_{q,e}
=
\mathcal{Z}_{q,e}^{\mathcal{A}_q}
\times
\left(
\mathcal{Z}_{q,e}
\setminus
\mathcal{Z}_{q,e}^{\mathcal{A}_q}
\right),
\end{equation}
and optimize
\begin{equation}
\mathcal{L}_{\mathrm{rank}}^{q,e}
=
\frac{1}{|\mathcal{P}_{q,e}|}
\sum_{(z^+,z^-)\in\mathcal{P}_{q,e}}
\left[
1-s_\phi(q,z^+)+s_\phi(q,z^-)
\right]_+,
\label{eq:ablation_rank}
\end{equation}
where $[x]_+=\max(0,x)$. We use the standard margin of $1$ and enumerate the full Cartesian product without pair sampling. Question--entity pairs without both positive and negative paths are excluded. For both baselines, the per-pair losses are averaged across the batch.

Both baselines use the same reward-model architecture, candidate paths, optimizer, learning rate, batch size, and training epochs as \framework.

The \textit{w/o question} variant removes the question representation from the reward model while keeping the asymmetric reward-learning objective and all other settings unchanged.

\textbf{Transferability Settings.}
We further evaluate whether the learned path rewards can supervise other KG retrieval architectures. We consider RoG~\citep{luo_rog_2024}, which generates relation paths with an LLM, and SubgraphRAG~\citep{li_subgraphrag_2025}, which trains a lightweight triple retriever.

For \textit{RoG + \framework Reward Sup.}, we retain the RoG architecture and use the highest-scoring answer-reaching path for each question--entity pair as training supervision:
\begin{equation}
z^{*}_{q,e}
=
\arg\max_{z\in\mathcal{Z}_{q,e}^{\mathcal{A}_q}}
s_\phi(q,z).
\label{eq:rog_reward_sup}
\end{equation}

For \textit{SubgraphRAG + \framework Reward Sup.}, we retain its original shortest-path supervision and reweight each supervised path using the learned reward. For a path $z$ with reward score $c=s_\phi(q,z)$, we assign
\begin{equation}
w(z)
=
\exp(c-c_{\max}),
\qquad
c_{\max}
=
\max_{z'\in\mathcal{Z}_{q,e}}
s_\phi(q,z'),
\label{eq:subgraphrag_reward_weight}
\end{equation}
where $c_{\max}$ is computed over the forward paths scored for the same question--entity pair. This is equivalent to the temperature-$1$ softmax score normalized by its maximum. All other training and answer-generation settings follow the corresponding original baselines.

\textbf{Reward-Guided Path Generation Variants.}
We consider five variants to examine the two stages of reward-guided path generation.

\textit{Distillation only} retains path reward distillation and removes GRPO.

\textit{GRPO only} removes path reward distillation and performs GRPO with the learned path reward directly on the model.

\textit{Distillation + GRPO-Ans} retains path reward distillation but replaces the learned reward in GRPO with the binary answer-reaching reward
\begin{equation}
u_i^{\mathrm{ans}}
=
\mathbb{I}
\left[
\mathcal{E}_{\tilde{z}_i}(e)
\cap
\mathcal{A}_q
\neq
\emptyset
\right].
\label{eq:ablation_answer_reward}
\end{equation}

\textit{Distillation + GRPO-F1} instead uses the answer-set F1 of the retrieved entities as the GRPO reward,
\begin{equation}
u_i^{\mathrm{F1}}
=
\mathrm{F1}\!\left(\mathcal{E}_{\tilde{z}_i}(e),\,\mathcal{A}_q\right).
\label{eq:ablation_f1_reward}
\end{equation}

For both variants, the same output-format, hop-length, and KG-grounding validity constraints as the full model are applied before reward assignment, and all other GRPO settings remain unchanged.

Finally, \textit{SFT + GRPO-Ans} removes the learned reward from both stages. The first stage performs uniform SFT over answer-reaching paths,
\begin{equation}
p_{\mathrm{uni}}(z_i\mid q,e)
=
\frac{1}{|\mathcal{Z}_{q,e}^{\mathcal{A}_q}|},
\qquad
\mathcal{L}_{\mathrm{uniform}}
=
-\sum_{z_i\in\mathcal{Z}_{q,e}^{\mathcal{A}_q}}
p_{\mathrm{uni}}(z_i\mid q,e)
\log\pi_\theta(z_i\mid q,e),
\label{eq:ablation_uniform}
\end{equation}
and the second stage performs GRPO with the binary answer reward in Eq.~\ref{eq:ablation_answer_reward}. This variant corresponds to the standard SFT-then-GRPO recipe.

\textbf{Results.}
The complete results in Table~\ref{tab:full_ablation} are consistent with Section~\ref{sec:ablation}. On Hits@1, the asymmetric, question-conditioned reward again outperforms both conventional objectives, and its benefits transfer to RoG and SubgraphRAG. For path generation, distillation provides the initialization, whereas GRPO with the binary answer reward or answer-set F1 brings no consistent gain over distillation alone on Hits@1 either. Only GRPO with the learned reward improves all three metrics, and the standard SFT-then-GRPO recipe remains the weakest distilled variant.

\subsection{Additional SPARQL Path Ranking Results}\label{app:additional_path_ranking}

\begin{table}[h]
\centering
\small
\caption{SPARQL path ranking performance (\%) on the development sets of WebQSP and CWQ.}
\label{tab:reward_quality_dev}
\begin{tabular}{lcccc}
\toprule
\multirow[b]{2}{*}{\textbf{Scoring Method}}
& \multicolumn{2}{c}{\textbf{WebQSP}}
& \multicolumn{2}{c}{\textbf{CWQ}} \\
\cmidrule(lr){2-3} \cmidrule(lr){4-5}
& \textbf{Hits@1} & \textbf{MRR}
& \textbf{Hits@1} & \textbf{MRR} \\
\midrule
Cosine sim.
& 58.6 & 71.6 & 26.7 & 42.7 \\
Answer F1
& 75.9 & 84.9 & 21.3 & 38.2 \\
Pointwise BCE
& 69.8 & 79.1 & 33.9 & 46.2 \\
Pairwise ranking
& 69.1 & 78.9 & 37.2 & 48.4 \\
GPT-4o
& \uline{83.3} & \uline{88.2} & \uline{59.7} & \uline{69.4} \\
\midrule
\framework Reward
& \textbf{88.3} & \textbf{91.9}
& \textbf{70.3} & \textbf{78.0} \\
\bottomrule
\end{tabular}
\end{table}

Table~\ref{tab:reward_quality_dev} reports SPARQL path ranking results on the development sets, using the full WebQSP development set and 200 randomly sampled CWQ development questions with more than one answer-reaching candidate path, selected as in Section~\ref{sec:reward_analysis}. The results are consistent with the main analysis: \framework Reward achieves the best hits@1 and MRR on both datasets, leading GPT-4o by 5.0 and 10.6 Hits@1 points on WebQSP and CWQ and the remaining scorers by larger margins.

\subsection{Results on MetaQA}\label{app:metaqa}

\begin{table}[h]
\centering
\small
\caption{KGQA performance (\%) on MetaQA. $^\dagger$ denotes a fine-tuned LLM answer reasoner, and $^*$ denotes results reported in the original paper.}
\label{tab:metaqa}
\begin{tabular}{llccc}
\toprule
\textbf{Method} & \textbf{LLM}
& \textbf{F1} & \textbf{Hit} & \textbf{Hits@1} \\
\midrule
ToG~\citep{sun_tog_2024}
& GPT-4o
& 86.3 & 87.7 & 80.1 \\

DoG~\citep{ma_debate_2025}
& GPT-4o
& 93.3 & 98.3 & 90.4 \\

RoG~\citep{luo_rog_2024}
& LLaMA2-Chat-7B$^\dagger$
& 88.1 & 92.7 & 90.3 \\

DP*~\citep{ma_dp_2025}
& GPT-4o
& 94.4 & 96.5 & 95.2 \\

\midrule
\framework
& GPT-4o
& \textbf{96.2} & \textbf{99.8} & \textbf{97.5} \\
\bottomrule
\end{tabular}
\end{table}

To evaluate \framework on a different KG, we additionally conduct experiments on MetaQA~\citep{zhang_metaqa_2018}, a 1- to 3-hop KGQA benchmark built on WikiMovies. Following prior work~\citep{ma_dp_2025, ma_debate_2025}, we evaluate on 200 questions for each hop.

As shown in Table~\ref{tab:metaqa}, \framework achieves the best F1, Hit, and Hits@1 among all methods. The effectiveness of reward-guided path generation thus extends to a different KG.

\section{Parameter Study}\label{app:param_study}

\begin{figure}[h]
    \centering
    \begin{subfigure}[t]{0.48\textwidth}
        \centering
        \includegraphics[width=\linewidth]{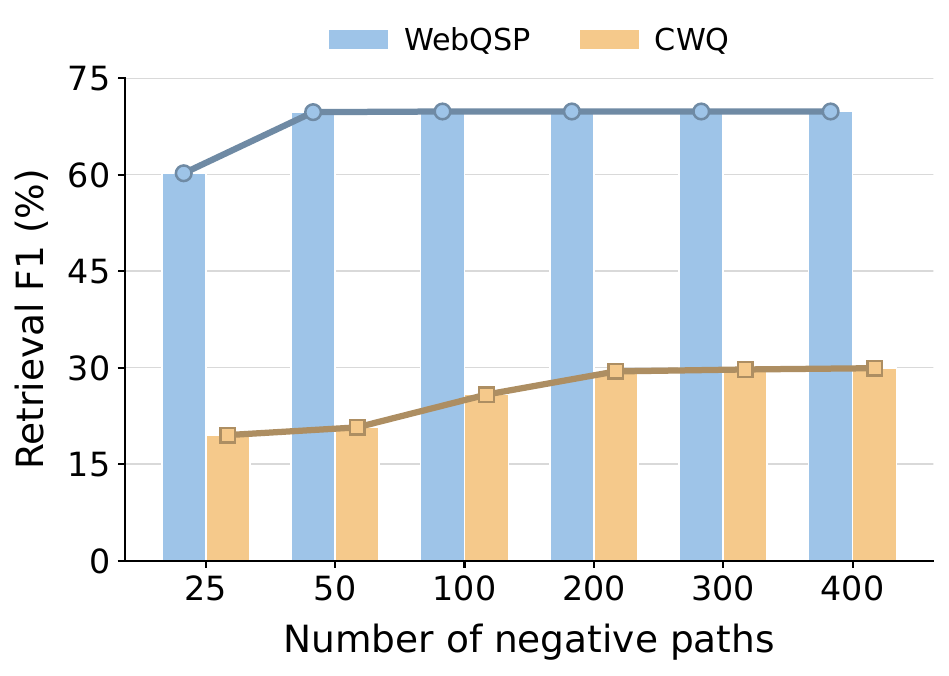}
        \caption{Effect of the number of negative paths on grounded entity retrieval F1.}
        \label{fig:negative_paths}
    \end{subfigure}
    \hfill
    \begin{subfigure}[t]{0.48\textwidth}
        \centering
        \includegraphics[width=\linewidth]{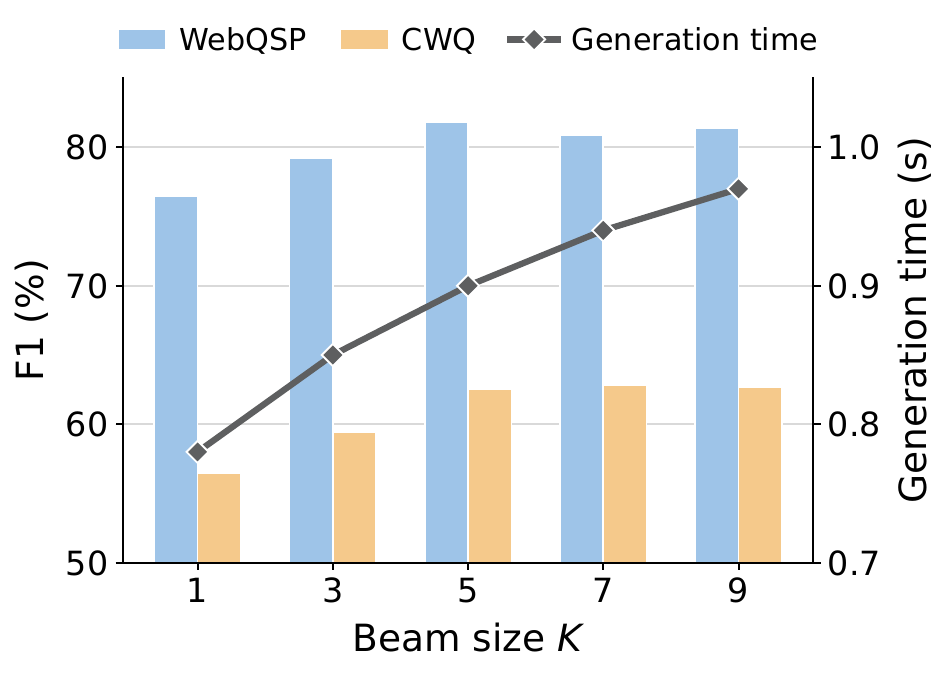}
        \caption{Effect of beam size on answer F1 and generation runtime.}
        \label{fig:beam_size}
    \end{subfigure}
    \caption{Parameter study of \framework.}
    \label{fig:parameter_study}
\end{figure}

In this section, we conduct the parameter study on the development sets of WebQSP and CWQ. Figure~\ref{fig:parameter_study} summarizes the sensitivity of \framework to the number of sampled negative paths during reward-model training and the beam size used at inference.

\textbf{Number of Negative Paths.}
As shown in Figure~\ref{fig:negative_paths}, increasing the number of sampled negative paths improves grounded entity retrieval F1, with the gains saturating after 50 negatives on WebQSP and around 200 negatives on CWQ. We therefore use 50 and 200 negative paths for WebQSP and CWQ, respectively, which provide a favorable trade-off between retrieval quality and training cost.

\textbf{Beam Size.}
Figure~\ref{fig:beam_size} shows that increasing the beam size improves answer F1 initially, while also increasing generation time. Performance largely saturates beyond $K=5$, with only marginal changes at larger beam sizes. We therefore set $K=5$ for all experiments to balance answer quality and inference efficiency.

\section{Case Study}\label{app:case_study}

We provide two examples to illustrate how the learned path rewards distinguish ambiguous answer-reaching paths and how reward-guided path generation affects downstream KG-grounded answer reasoning.

\textbf{Path Reward Learning.}

\begin{wrapfigure}{r}{0.36\textwidth}
    \vspace{-1.2em}
    \centering
    \includegraphics[width=\linewidth]{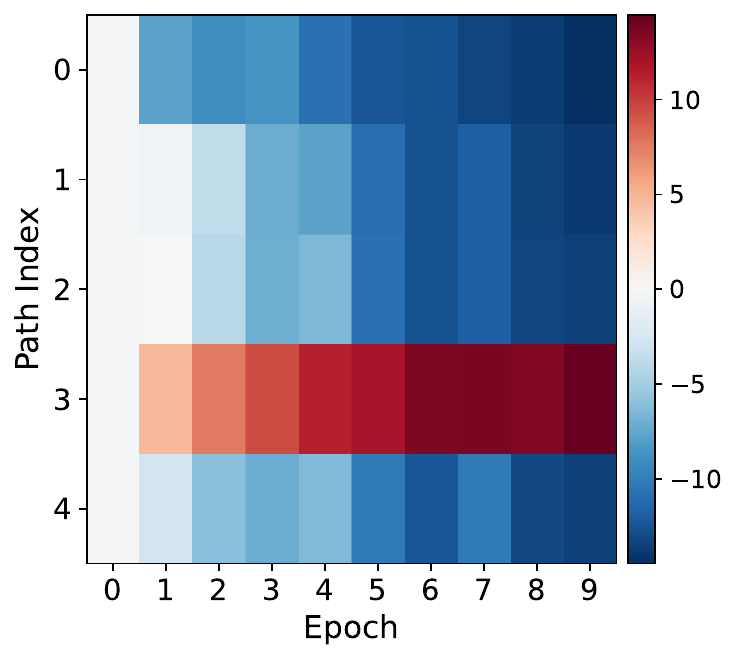}
    \caption{Evolution of learned reward scores for candidate paths in the example from Table~\ref{tab:case_study_supervision}.}
    \label{fig:case_study_reward}
    \vspace{-1.0em}
\end{wrapfigure}

Table~\ref{tab:case_study_supervision} presents a CWQ training example in which multiple candidate paths reach plausible answers. The question asks for the language spoken in the region where the \textit{Khorshid} newspaper is circulated. Path~0 directly retrieves the newspaper language but bypasses the circulation-region constraint in the question. In contrast, Path~3 first reaches the newspaper's circulation region and then retrieves the languages spoken there, matching the annotated SPARQL relation path.

\begin{table}[t]
    \centering
    \small
    \caption{Example of path reward learning on a CWQ training question. Green denotes the SPARQL-consistent path, while red denotes a spurious answer-reaching path.}
    \label{tab:case_study_supervision}
    \resizebox{\linewidth}{!}{
    \begin{tabular}{l}
        \toprule
        \textbf{Question:} What language would be spoken in the region where the Khorshid newspaper is circulated? \\
        \textbf{Question Entity:} Khorshid \\
        \midrule

        \cellcolor{gray!12}\textbf{\textit{Candidate Answer-Reaching Paths:}} \\
        \quad\textbf{Path 0:} book.periodical.language \\
        \quad\textbf{Path 1:} book.periodical.language
        $\rightarrow$language.human\_language.countries\_spoken\_in
        $\rightarrow$location.country.official\_language \\
        \quad\textbf{Path 2:} book.newspaper.circulation\_areas
        $\rightarrow$film.film\_subject.films
        $\rightarrow$film.film.language \\
        \quad\textbf{Path 3:} book.newspaper.circulation\_areas
        $\rightarrow$location.country.languages\_spoken \\
        \quad\textbf{Path 4:} book.periodical.language
        $\rightarrow$language.human\_language.language\_family
        $\rightarrow$language.language\_family.languages \\
        \midrule

        \textbf{Top-ranked Path by RAPL (GPT-4o):}
        \textcolor{red}{book.periodical.language} \\
        \textbf{Top-ranked Path by \framework Reward:}
        \textcolor{green!60!black}{
        book.newspaper.circulation\_areas
        $\rightarrow$location.country.languages\_spoken} \\
        \bottomrule
    \end{tabular}
    }
\end{table}

RAPL ranks the shortcut Path~0 highest, whereas the learned \framework reward assigns the highest score to the SPARQL-consistent Path~3. Figure~\ref{fig:case_study_reward} further shows how the reward-induced preferences over the candidate paths evolve during training. The preference gradually shifts toward Path~3, illustrating how the asymmetric objective can differentiate paths that are indistinguishable under binary answer-reaching supervision.

\textbf{KG-Grounded Answer Reasoning.}
Table~\ref{tab:case_study_reasoning} illustrates how path supervision affects downstream KG-grounded answer reasoning. For the WebQSP question ``Who were Demeter's siblings?'', the variant without learned path rewards generates a spurious parent--children path and an incomplete sibling path, resulting in unresolved Freebase entities in the final prediction. In contrast, \framework generates a grounded relation path that reaches the correct answer, Zeus.

\begin{table}[h]
    \centering
    \small
    \caption{Example of KG-grounded answer reasoning on a WebQSP test question.}
    \label{tab:case_study_reasoning}
    \resizebox{\linewidth}{!}{
    \begin{tabular}{cl}
        \toprule
        \textbf{Question} & Who were Demeter's siblings? \\
        \textbf{Answer} & Zeus \\
        \midrule

        \multirow{10}{*}{\textbf{\makecell{SFT + \\GRPO-Ans}}}
        & \cellcolor{gray!12}\textit{\textbf{Generated Paths:}} \\
        & [PATH 1] \\
        &\quad Start Entity: Demeter \\
        &\quad Path: fictional\_universe.fictional\_character.parents
        $\rightarrow$fictional\_universe.fictional\_character.children \\
        &\quad End Entities: Hera, Poseidon, Hestia, Hades, Zeus \\
        & [PATH 2] \\
        &\quad Start Entity: Demeter \\
        &\quad Path: fictional\_universe.fictional\_character.siblings \\
        &\quad End Entities: m.0gwhv5j, m.0j85m5t \\
        \cmidrule{2-2}
        & \cellcolor{gray!12}\textit{\textbf{Predicted Answer:}} \\
        &\quad \textcolor{red}{m.0gwhv5j, m.0j85m5t} \\

        \midrule

        \multirow{8}{*}{\textbf{\framework}}
        & \cellcolor{gray!12}\textit{\textbf{Generated Paths:}} \\
        & [PATH 1] \\
        &\quad Start Entity: Demeter \\
        &\quad Path: fictional\_universe.fictional\_character.siblings
        $\rightarrow$ \\
        &\quad\quad fictional\_universe.sibling\_relationship\_of\_fictional\_characters.siblings \\
        &\quad End Entities: Zeus \\
        \cmidrule{2-2}
        & \cellcolor{gray!12}\textit{\textbf{Predicted Answer:}} \\
        &\quad \textcolor{green!60!black}{Zeus} \\
        \bottomrule
    \end{tabular}
    }
\end{table}

This example complements the quantitative ablations: the learned path reward helps the generator avoid shortcut or incomplete relation paths and provides KG-grounded evidence that reflects the reasoning required by the question.

\section{Error Analysis}\label{app:error_analysis}

\begin{wrapfigure}[17]{r}{0.4\textwidth} 
\vspace{-1.5em} 
\centering
\includegraphics[width=\linewidth]{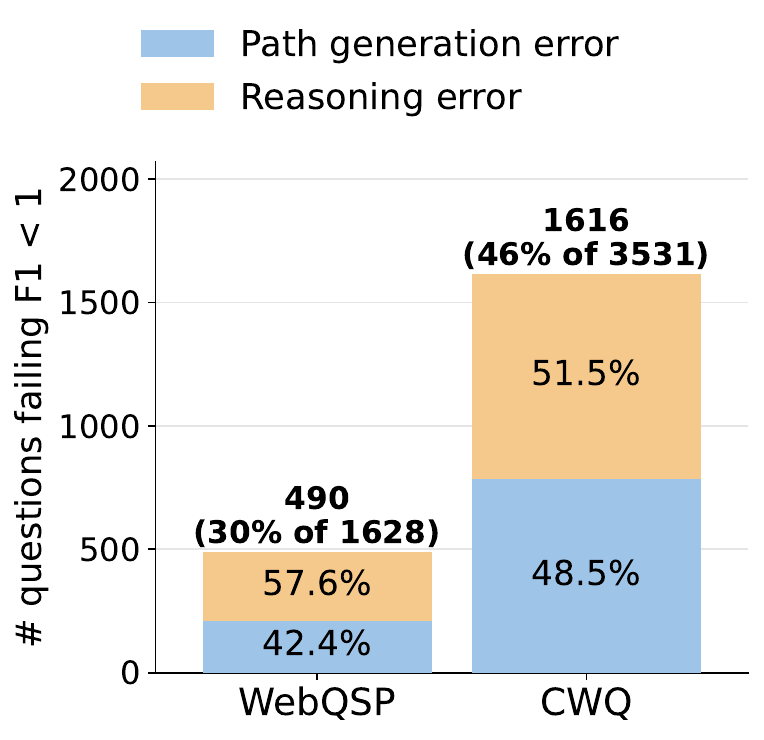}
\caption{Breakdown of error types among cases with F1 $<1$ on WebQSP and CWQ.} 
\label{fig:error_analysis} 
\end{wrapfigure}

We analyze test questions with F1 $<1$ and categorize the failures into path generation errors, where none of the generated paths matches the SPARQL reference path, and reasoning errors, where reference paths are generated but the answer LLM fails to return the correct answers. As shown in Figure~\ref{fig:error_analysis}, 490 WebQSP questions (30\%) and 1,616 CWQ questions (46\%) fall into this category. On WebQSP, reasoning errors account for 57.6\% of the failures and path generation errors for 42.4\%, while on CWQ the two sources are more evenly split, at 51.5\% and 48.5\%. Neither source dominates on either dataset. Path generation errors propagate by producing incomplete or irrelevant evidence for answer reasoning, whereas reasoning errors arise even when the reference path is retrieved, as the answer LLM fails to identify or compose the correct answers. We discuss the implications in Appendix~\ref{app:limitation}.

\section{Limitations}\label{app:limitation}

\framework follows a staged pipeline of relation path generation, KG grounding, and answer reasoning. As shown in Appendix~\ref{app:error_analysis}, failures occur in both path generation and answer reasoning, and errors in an earlier stage propagate to the final prediction. A unified agentic framework that coordinates path generation, evidence retrieval, and answer reasoning, allowing feedback across stages, may mitigate such errors. Second, \framework relies on the KG to provide sufficient evidence. When relevant facts or connecting paths are missing, the generated paths cannot recover the information needed for reasoning, a limitation shared by all KG-grounded approaches.

\end{document}